\documentclass[11pt]{article}

\usepackage{graphicx}%
\usepackage{multirow}%
\usepackage{amsmath,amssymb,amsfonts}%
\usepackage{amsthm}%
\usepackage{mathrsfs}%
\usepackage[title]{appendix}%
\usepackage{xcolor}%
\usepackage{textcomp}%
\usepackage{manyfoot}%
\usepackage{booktabs}%
\usepackage{algorithm}%
\usepackage{algorithmicx}%
\usepackage{algpseudocode}%
\usepackage{listings}%
\usepackage{pdfpages}
\usepackage[letterpaper,top=2cm,bottom=2cm,left=3cm,right=3cm,marginparwidth=1.75cm]{geometry}
\usepackage{bigstrut}
\usepackage{xspace}
\usepackage{longtable}
\usepackage{caption}
\usepackage{subcaption}
\usepackage{adjustbox}
\usepackage[colorlinks=true, linkcolor=black, citecolor=blue]{hyperref}
\usepackage[superscript]{cite}
\usepackage{lineno}

\makeatletter
\renewcommand{\maketitle}{\bgroup\setlength{\parindent}{0pt}
\begin{flushleft}
  \textbf{\@title}

  \@author
\end{flushleft}\egroup
}
\makeatother

\begin{document}

\newcommand{\method}{\texttt{TrialMind}\xspace}
\newcommand{\dataset}{\texttt{TrialReviewBench}\xspace}
\newcommand{\recall}{\text{Recall}\xspace}
\newcommand{\acc}{\text{ACC}\xspace}
\newcommand{\precision}{\text{Precision}\xspace}
\newcommand{\llm}{\text{LLM}\xspace}

\title{Accelerating Clinical Evidence Synthesis with Large Language Models}
\author{Zifeng Wang$^{1}$, Lang Cao$^1$, Benjamin Danek$^1$, Qiao Jin$^{2}$, Zhiyong Lu$^{2}$, Jimeng Sun$^{1,3\#}$\\
$^1$ Department of Computer Science, University of Illinois Urbana-Champaign, Champaign, IL \\
$^2$ National Center for Biotechnology Information, National Library of Medicine, Bethesda, MD  \\
$^3$ Carle Illinois College of Medicine, University of Illinois Urbana-Champaign, Champaign, IL \\
$^\#$Corresponding authors. Emails: jimeng@illinois.edu}
\maketitle



\abstract{
Synthesizing clinical evidence largely relies on systematic reviews of clinical trials and retrospective analyses from medical literature. However, the rapid expansion of publications presents challenges in efficiently identifying, summarizing, and updating clinical evidence. Here, we introduce \method, a generative artificial intelligence (AI) pipeline for facilitating human-AI collaboration in three crucial tasks for evidence synthesis: study search, screening, and data extraction. To assess its performance, we chose published systematic reviews to build the benchmark dataset, named \dataset, which contains 100 systematic reviews and the associated 2,220 clinical studies. Our results show that \method excels across all three tasks. In study search, it generates diverse and comprehensive search queries to achieve high recall rates (Ours 0.711-0.834 v.s. Human baseline 0.138-0.232). For study screening, \method surpasses traditional embedding-based methods by 30\% to 160\%. In data extraction, it outperforms a GPT-4 baseline by 29.6\% to 61.5\%. We further conducted user studies to confirm its practical utility. Compared to manual efforts, human-AI collaboration using \method yielded a 71.4\% recall lift and 44.2\% time savings in study screening and a 23.5\% accuracy lift and 63.4\% time savings in data extraction. Additionally, when comparing synthesized clinical evidence presented in forest plots, medical experts favored \method's outputs over GPT-4's outputs in 62.5\% to 100\% of cases. These findings show the promise of LLM-based approaches like \method to accelerate clinical evidence synthesis via streamlining study search, screening, and data extraction from medical literature, with exceptional performance improvement when working with human experts.
}

\newpage

\section*{Introduction}\label{sec:intro}
Clinical evidence is crucial for supporting clinical practices and advancing new drug development and needs to be updated regularly~\cite{elliott2021decision}. It is primarily gathered through retrospective analysis of real-world data or through prospective clinical trials that assess new interventions on humans. Researchers usually conduct systematic reviews to consolidate evidence from various clinical studies in the literature~\cite{field2010meta,concato2017randomized}. However, this process is expensive and time-consuming, requiring an average of five experts and 67.3 weeks based on an analysis of 195 systematic reviews~\cite{borah2017analysis}. Moreover, the fast growth of clinical study databases means that the information in these published clinical reviews becomes outdated rapidly~\cite{hoffmeyer2021most}. For instance, PubMed has indexed over 35M citations and gets over 1M new citations annually~\cite{pubmednum}. This situation underscores the urgent need to streamline the systematic review processes to document systematic and timely clinical evidence from the extensive medical literature~\cite{marshall2019toward, elliott2021decision}.

Large language models (LLMs) excel at information processing and generating. They can be adapted to target tasks by providing the task definition and examples as the inputs (namely ``prompts'')~\cite{brown2020language}. Researchers have tried to adopt LLMs for many individual tasks in the evidence synthesis process, including generating searching queries~\cite{wang2023can,adam2024literature}, extracting studies' population, intervention, comparison, outcome (PICO) elements~\cite{wadhwa2023jointly,zhang2024span}, screening citations~\cite{syriani2023assessing}, and summarizing findings from multiple studies~\cite{shaib2023summarizing, wallace2021generating,zhang2024closing,peng2023ai}. However, few have investigated LLMs' effectiveness across the entire evidence synthesis process~\cite{christopoulou2023towards}. This is crucial because it ensures a seamless integration of AI in every step, potentially improving overall efficiency and accuracy. Understanding the strengths and limitations of LLMs in a holistic manner enables more effective automation and human-AI collaboration. To fill this gap, we created a testing dataset \dataset that covers major tasks in evidence synthesis, including study search, screening, and data extraction tasks. We chose published systematic reviews to create the dataset. As a result, the dataset includes 100 systematic reviews with 2,220 associated clinical studies. It also consists of manual annotations of 1,334 study characteristics and 1,049 study results. Based on \dataset, we are able to assess cutting-edge LLMs, e.g., GPT-4~\cite{openai2024gpt4}, in clinical evidence synthesis tasks.

\begin{figure}[htbp]
    \centering
    \includegraphics[width=\linewidth]{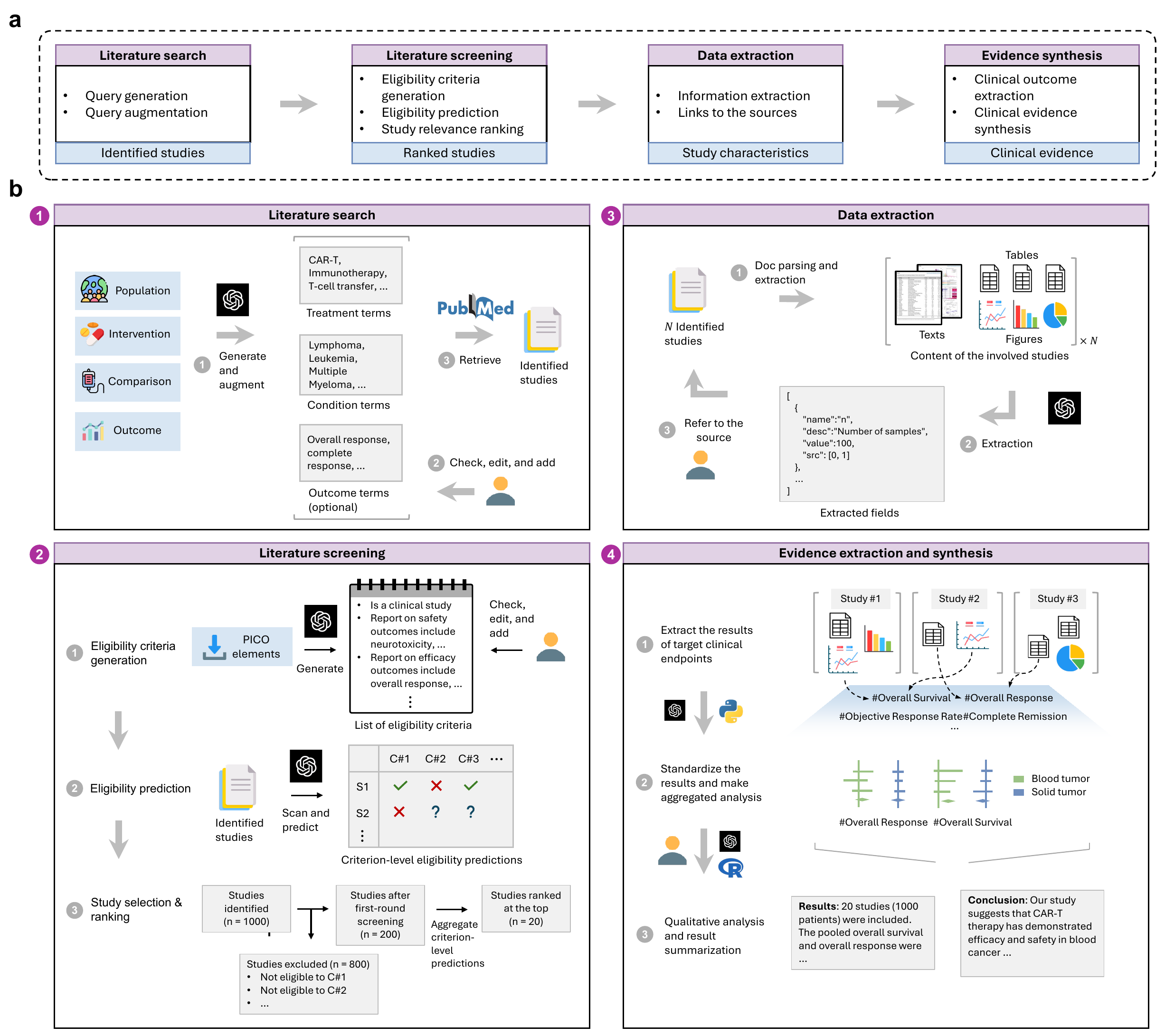}
    \caption{\textbf{The overview of \method pipeline. a}, it has four main steps: literature search, literature screening, data extraction, and evidence synthesis. \textbf{b}, (1) Utilizing input PICO elements, \method generates key terms to construct Boolean queries for retrieving studies from literature databases. (2) \method formulates eligibility criteria, which users can edit to provide context for LLMs during eligibility predictions. Users can then select studies based on these predictions and rank their relevance by aggregating them. (3) \method processes the descriptions of target data fields to extract and output the required information as structured data. (4) \method extracts findings from the studies and collaborates with users to synthesize the clinical evidence.}
\label{fig:overview}
\end{figure}

Furthermore, this study aims to fill the gap in adapting LLMs to evidence synthesis tasks, overcoming LLM's limitations in (1) hallucinations, (2) weakness in reasoning with numerical data, (3) overly generic outputs, and (4) lack of transparency and reliability~\cite{yun2023appraising}. Specifically, we developed an AI-driven pipeline named \method, which is optimized for (1) generating boolean queries to search citations from the literature; (2) building eligibility criteria and screening through the found citations; and (3) extracting data, including study protocols, methods, participant baselines, study results, etc., from publications and reports. More importantly, \method breaks down into subtasks that adhere to the established practice of systematic reviews~\cite{page2021prisma}, which facilitates experts in the loop to monitor, edit, and verify intermediate outputs. It also has the flexibility to allow experts to begin at any intermediate step as needed.

In this study, we show that the \method is able to 1) retrieve a complete list of target studies from the literature, 2) follow the specified eligibility criteria to rank the most relevant studies at the top, and 3) achieve high accuracy in extracting information and clinical outcomes from unstructured documents based on user requests. Beyond providing descriptive evidence, \method can extract numerical clinical outcomes to be standardized as input for meta-analysis (e.g., forest plots). A human evaluation was conducted to assess the synthesized evidence. Finally, to validate the practical benefits, we developed an accessible web application based on \method and conducted a user study comparing two approaches: AI-assisted experts versus standalone experts. We measured the time savings and evaluated the output quality of each approach. The results show that \method significantly reduced the time required for study search, citation screening, and data extraction, while maintaining or improving the quality of the output compared to experts working alone.

\section*{Results}\label{sec:results}
\subsection*{Creating \dataset from medical literature}
\begin{figure}[htbp]
    \centering
    \includegraphics[width=\linewidth]{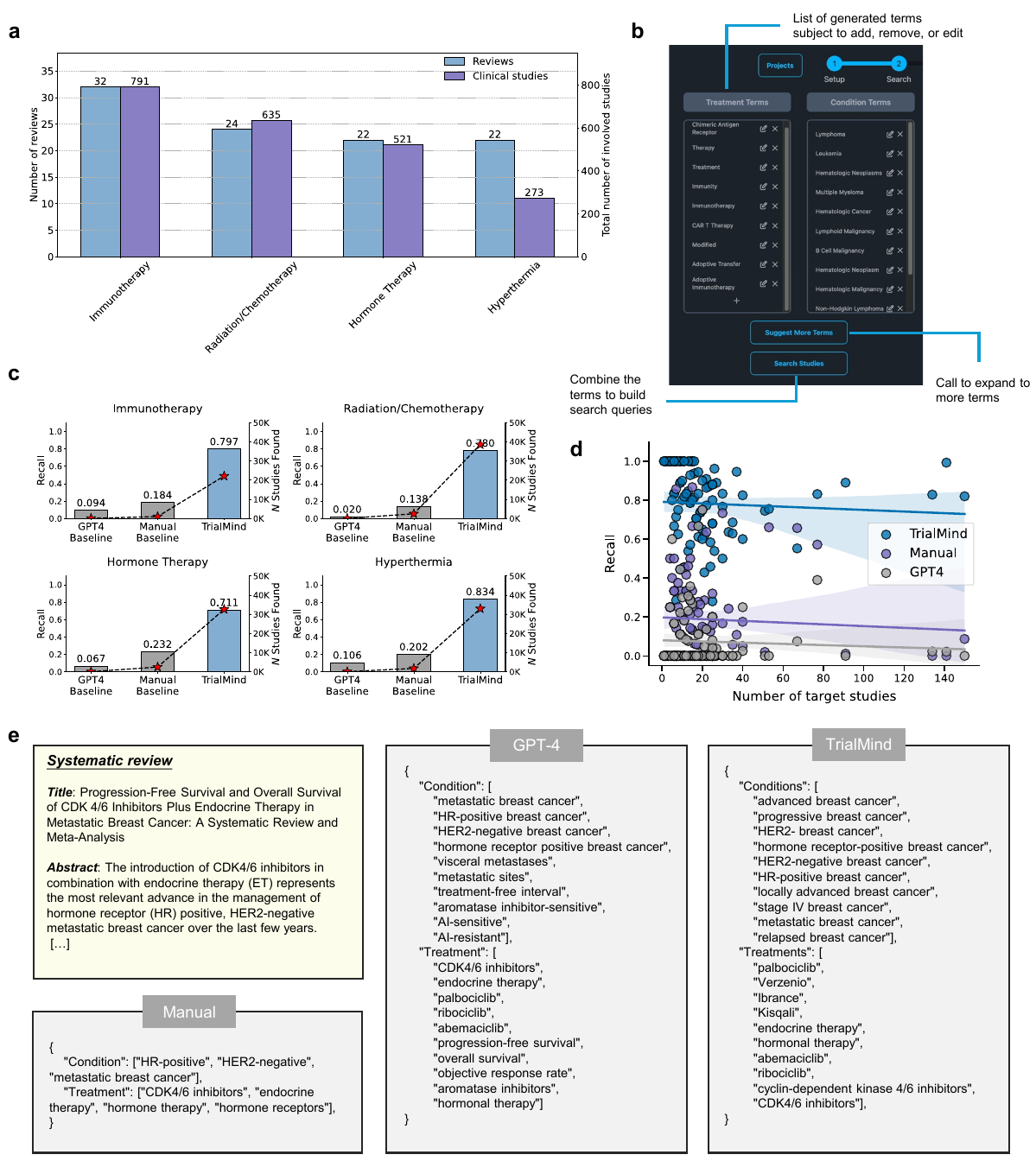}
    \caption{\textbf{Literature search experiment results.} \textbf{a}, The total number of involved studies and the number of review papers across different topics. \textbf{b}, The \method's interface for users to retrieve studies. \textbf{c}, the Recall of the search results for reviews across four topics. The bar heights indicate the Recall, and the star indicates the number of studies found. \textbf{d}, Scatter plots of the Recall against the number of ground-truth studies. Each scatter indicates the results of one review. Regression estimates are displayed with the 95\% CIs in blue or purple. \textbf{e}, Example cases comparing the outputs of three methods.}
    \label{fig:exp_search_recall_combined}
\end{figure}

A systematic understanding of cancer treatments is crucial for oncology drug discovery and development. We retrieved a list of cancer treatments from the National Cancer Institute's introductory page as the keywords to search medical systematic reviews~\cite{ncicancer}. To ensure data quality, we crafted comprehensive queries with automatic filtering and manual screening. For each review, we obtained the list of studies with their PubMed IDs, retrieved their full content, and extracted study characteristics and clinical outcomes. We followed PubMed's usage policy and guidelines during retrieval. Further manual checks were performed to correct inaccuracies, eliminate invalid and duplicate papers, and refine the text for clarity (Methods). The final \dataset dataset consists of 2,220 studies involved in 100 reviews (Fig.~\hyperref[fig:exp_search_recall_combined]{2a}), covering four major topics: Immunotherapy, Radiation/Chemotherapy, Hormone Therapy, and Hyperthermia. We manually created three major evaluation tasks based on these reviews: study search, study screening, and data extraction.

The study search task begins with the PICO (Population, Intervention, Comparison, Outcome) elements extracted from the abstract of a systematic review, which serve as the formal definition of the research question. The model being tested is tasked with generating relevant keywords for the treatment and condition terms, as depicted in Fig.~\hyperref[fig:exp_search_recall_combined]{2e}. These keywords are then used to form Boolean queries, which are submitted to search citations in the PubMed database. The performance of the model is evaluated by checking whether the retrieved studies include those that were actually involved in the target systematic review. The recall rate is computed by measuring the proportion of actually involved studies identified through the search.

For the study screening task, the input consists of the PICO elements defined in the target systematic review. A candidate set of 2,000 citations is created by combining the actual studies included in the review with additional citations retrieved during the search but not included in the review. The model being tested ranks these citations based on the likelihood that each citation should be included in the systematic review. To assess the model's performance, we compute Recall@$k$: the recall value indicating how many of the actual included studies appear in the top $k$ ranked candidates.

The data extraction task focuses on retrieving specific information from the input study documents. In this case, we extract Table 1 from each systematic review, which typically details study characteristics such as study design, population demographics, and outcome measurements. These characteristics are matched to the individual studies and manually verified, yielding 1,334 study characteristic annotations. Additionally, we extract individual study results from the review’s reported analysis, often presented in forest plots, capturing metrics such as overall response and event rates, resulting in 1,049 study result annotations. The model being tested is given a list of target data points to extract, and its output is evaluated by assessing the accuracy of the extracted information based on the annotated datasets.

\subsection*{Build an LLM-driven system for clinical evidence synthesis}
Large language models (LLMs) excel in adapting to new tasks when provided with task-specific prompts while often struggling with complex tasks that require multiple steps of planning and reasoning. Additionally, interacting and collaborating with LLMs can be problematic due to their opaque nature and the complexity of debugging~\cite{wu2022ai}. In this study, we developed \method that decomposes the clinical evidence synthesis process into four main tasks (Fig.~\ref{fig:overview} and Methods). Initially, using the provided research question enriched with population, intervention, comparison, and outcome (PICO) elements, \method conducts a comprehensive search from the literature. It also works with users to build the eligibility criteria for target studies and then automate screening and ranking identified citations. Next, \method browses the study details to extract the study characteristics and pertinent findings. To ensure the accuracy and integrity of the data, each output is linked to the sources for manual inspection. In the final step, \method standardizes the clinical outcomes for meta-analysis.

\subsection*{\method can make a comprehensive retrieval of studies from the literature}
Finding relevant studies from medical literature like PubMed, which contains over 35 million entries, can be challenging. Typically, this requires the research expertise to craft complex queries that comprehensively cover pertinent studies. The challenge lies in balancing the specificity of queries: too stringent, and the search may miss relevant studies; too broad, and it becomes impractical to manually screen the overwhelming number of results. Previous approaches propose to prompt LLMs to generate the searching query directly~\cite{wang2023can}, which can induce incomplete searching results due to the limited knowledge of LLMs. In contrast, \method is designed to produce comprehensive queries through a pipeline that includes query generation, augmentation, and refinement. It also provides users with the ability to make further adjustments (Fig.~\hyperref[fig:exp_search_recall_combined]{2b}).

The dataset involving clinical studies spanning ten cancer treatment areas was used for evaluation (Fig.~\hyperref[fig:exp_search_recall_combined]{2a}). For each review, we collected the involved studies' PubMed IDs as the ground-truth and measured the Recall, i.e., how many ground-truth studies are found in the search results. We created two baselines as the comparison: GPT-4 and Human. The GPT-4 baseline makes a guided prompt for LLMs to generate the boolean queries~\cite{wang2023can}. It represents the common way of prompting LLMs for literature search query generation. The Human baseline represents a way where the key terms from PICO elements are extracted manually and expanded, referring to UMLS~\cite{bodenreider2004unified}, to construct the search queries. 

Overall, \method achieved a Recall of 0.782 on average for all reviews in \dataset, meaning it can capture most of the target studies. By contrast, the GPT-4 baseline yielded Recall = 0.073, and the Human baseline yielded Recall = 0.187. We divided the search results across four topics determined by the treatments studied in each review (Fig.~\hyperref[fig:exp_search_recall_combined]{2c}). Our analysis showed that \method can identify many more studies than the baselines. For instance, \method achieved $\recall = 0.797$ with identified studies $N$ = 22,084 for Immunotherapy-related reviews, while the GPT-4 baseline got $\recall = 0.094$ ($N$ studies = 27), and the Human baseline got $\text{Recall} = 0.154$ ($N$ studies = 958), respectively. In Radiation/Chemotherapy, \method achieved $\recall = 0.780$, the GPT-4 baseline got $\recall = 0.020$, and the Human baseline got $\recall = 0.138$. In Hormone Therapy, \method achieved $\recall = 0.711$, the GPT-4 baseline got $\recall = 0.067$, and the Human baseline got $\recall = 0.232$. In Hyperthermia, \method achieved $\recall = 0.834$, the GPT-4 baseline got $\recall = 0.106$, and the Human baseline got $\recall = 0.202$. These results demonstrate that regardless of the search task's complexity, as indicated by the variability in the Human baseline, \method consistently retrieves nearly all target studies from the PubMed database. This robust performance provides a solid foundation for accurately identifying target studies in the screening phase. 

Furthermore, we made scatter plots of Recall versus the number of target studies for each review (Fig.~\hyperref[fig:exp_search_recall_combined]{2d}). The hypothesis was that an increase in target studies correlates with the difficulty of achieving complete coverage. Our findings reveal that \method consistently maintained a high Recall, significantly outperforming the best baselines across all 100 reviews. A trend of declining Recall with an increasing number of target studies was confirmed through regression analysis. It was found that the GPT-4 baseline struggled, showing Recall close to 0, and the Human baseline results varied, with most reviews below 0.5. As the number of target studies increased, the Human and GPT-4 baselines' Recall decreased to nearly zero. In contrast, \method demonstrated remarkable resilience, showing minimal variation in performance despite the increasing number of target studies. For instance, in a review involving 141 studies, \method achieved a Recall of 0.99, while the GPT-4 and Human baselines obtained a Recall of 0.02 and 0, respectively.

\begin{figure}[htbp]
    \centering
    \includegraphics[width=\linewidth]{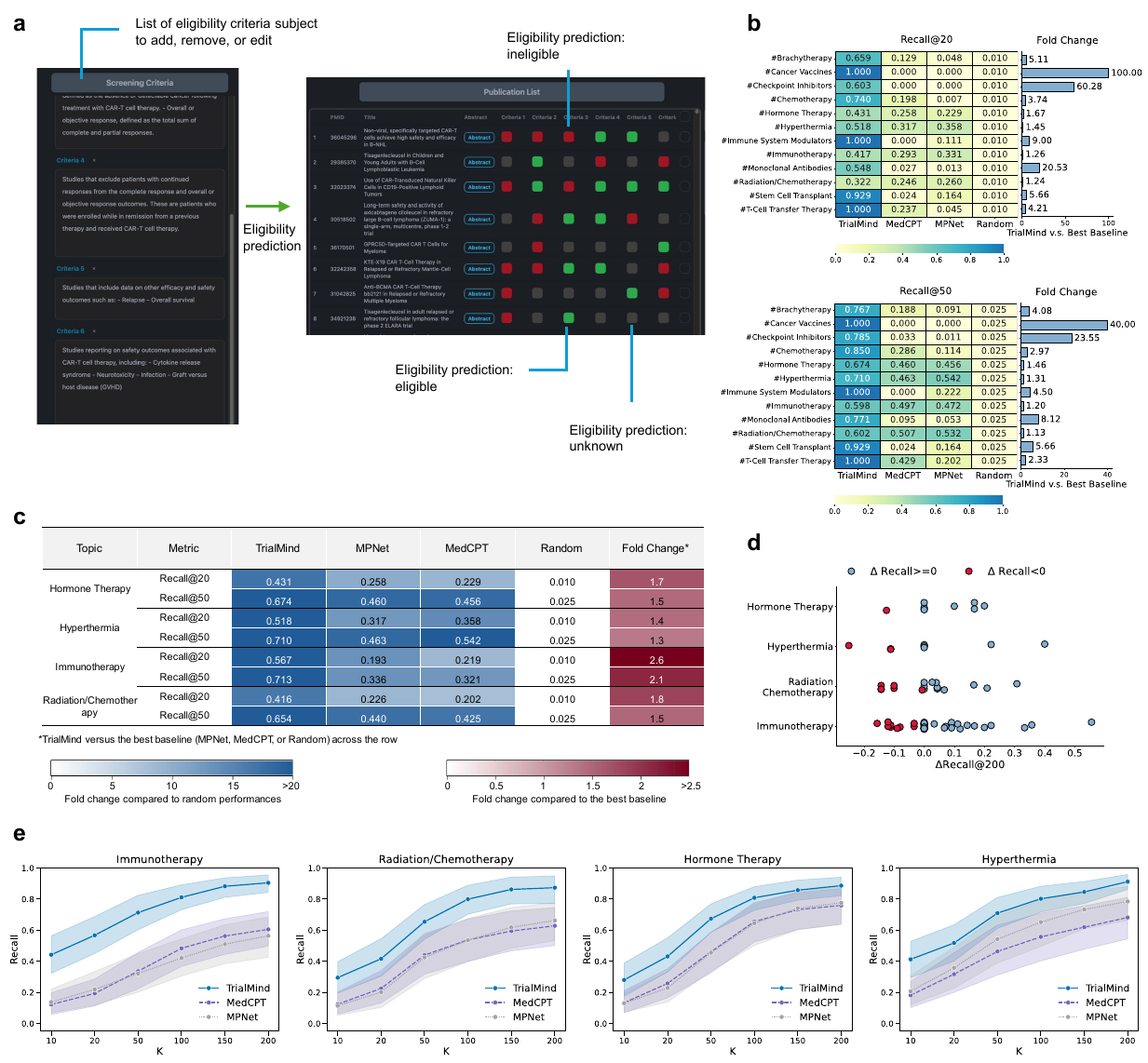}
    \caption{\textbf{Literature screen experiment results}. \textbf{a}, Streamline study screening using \method with human in the loop.  \textbf{b}, Ranking performances for Recall@20/50 within across therapeutic areas. \textbf{c}, Recall@20 and Recall@50 for \method and selected baselines. \textbf{d}, Effect of individual criterion on the ranking results. \textbf{e}, Ranking performance for $\recall@K$ with varying $K$ in four topics. Shaded areas are $95\%$ confidence interval.}
    \label{fig:exp_screen_combined}
\end{figure}

\subsection*{\method enhances literature screening and ranking}
Typically, human experts manually sift through thousands of retrieved studies to select relevant ones for inclusion in a systematic review. This process adheres to the PRISMA statement~\cite{page2021prisma}, which involves creating a list of eligibility criteria and assessing each study's eligibility. \method streamlines this task through a three-step approach: (1) it generates a set of inclusion criteria, which are subject to user's adjustments; (2) it applies these criteria to evaluate the study's eligibility, denoted by $\{-1,0,1\}$ where $-1$ and $1$ represent eligible and non-eligible, and $0$ represents unknown/uncertain, respectively; and (3) it ranks the studies by aggregating the eligibility predictions, where the aggregation strategy can be specified by users (Fig.~\hyperref[fig:exp_screen_combined]{3a}). We took a summation of the criteria-level eligibility predictions as the study-level relevance prediction scores for ranking. As such, \method provides a rationale for the relevance scores by detailing the eligibility predictions for each criterion.

We chose MPNet~\cite{song2020mpnet} and MedCPT~\cite{jin2023medcpt} as the general domain and medical domain ranking baselines, respectively. These methods compute study relevance by the cosine similarity between the encoded PICO elements as the query and the encoded study's abstracts. We also set a Random baseline that randomly samples from candidates. We created the evaluation data based on the search results in the first stage. For each review, we mixed the target studies with the other found studies to build a candidate set of 2,000 studies for ranking. Discriminating the target studies from the other candidates is challenging since all candidates meet the search queries, meaning they most probably investigate the relevant therapies or conditions. We evaluated the ranking performance using the Recall@20 and Recall@50 metrics. The concatenation of the title and abstract of each study is used for all methods as inputs.

We found that \method greatly improved ranking performances, with the fold changes over the best baselines ranging from 1.3 to 2.6 across four topics (Table~\hyperref[fig:exp_screen_combined]{3c}). For instance, for the Hormone Therapy topic, \method obtained $\recall@20 = 0.431$ and $\recall@50=0.674$. In the Hyperthermia topic, \method obtained $\recall@20 = 0.518$ and $\recall@50=0.710$. In the Immunotherapy topic, \method obtained $\recall@20 = 0.567$ and $\recall@50=0.713$. In the Radiation/Chemotherapy topic,  \method obtained $\recall@20 = 0.416$ and $\recall@50=0.654$. In contrast, other baselines exhibit significant variability across different topics. The general domain baseline MPNet was the worst as it performed similarly to the Random baseline in $\recall@20$. MedCPT showed marginal improvement over MPNet in the last three topics, while both failed to capture enough target studies in all topics.

Furthermore, \method demonstrated significant improvements over the baselines across various therapeutic areas (Fig.~\hyperref[fig:exp_screen_combined]{3b}). For example, in ``Cancer Vaccines" and ``Hormone Therapy," \method substantially increased $\recall@50$, achieving 33.33-fold and 10.53-fold improvements, respectively, compared to the best-performing baseline. \method generally attained a fold change greater than 2 (ranging from 1.57 to 33.33). Despite the challenge of selecting from a large pool of candidates ($n=2,000$) where candidates were very similar, \method identified an average of 43\% of target studies within the top 50. We compared \method to MedCPT and MPNet for $\recall@K$ ($K$ in 10 to 200) to gain insight into how $K$ influences the performances (Fig.~\hyperref[fig:exp_screen_combined]{3e}). We found \method can capture most of the target studies (over 80\%) when $K=100$. 

To thoroughly assess the quality of these criteria and their impact on ranking performance, we conducted a leave-one-out analysis to calculate $\Delta \recall@200$ for each criterion (Fig.~\hyperref[fig:exp_screen_combined]{3d}). The $\Delta \recall@200$ metric measures the difference in ranking performance with and without a specific criterion, with a larger value indicating superior criterion quality. Our findings revealed that most criteria positively influenced ranking performances, as the negative influence criteria are $n=1$ in Hormone Therapy, $n=1$ in Hyperthermia, $n=5$ in Radiation/Chemotherapy, and $n=7$ in Immunotherapy. Additionally, we identified redundancies among the generated criteria, as those with $\Delta \recall@200=0$ were the most frequently observed. This redundancy likely stems from some criteria covering similar eligibility aspects, thus not impacting performance when one is omitted.

\begin{figure}[htbp]
    \centering
    \includegraphics[width=0.88\linewidth]{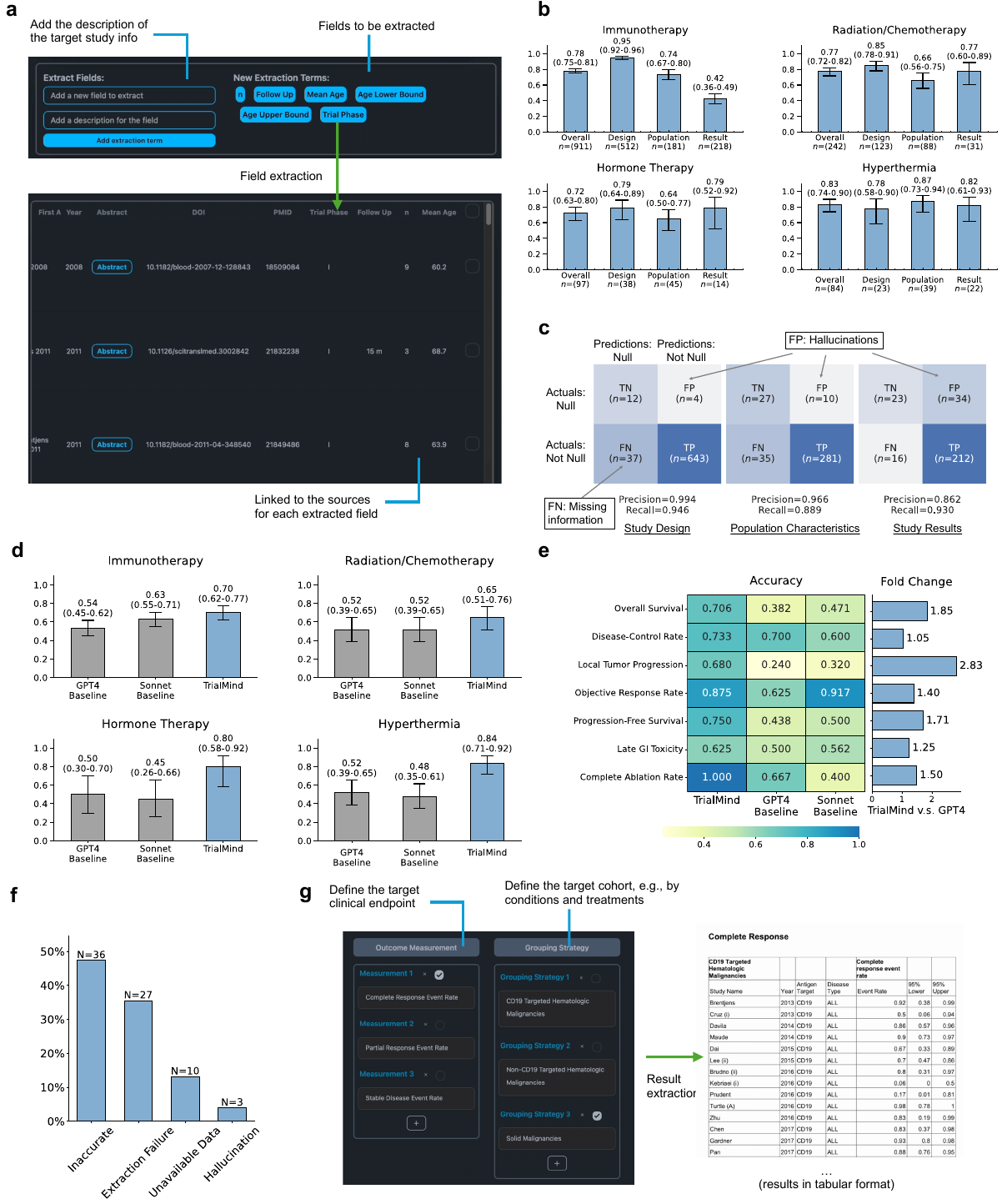}
    \caption{\textbf{Data and result extraction experiment results}. \textbf{a}, Streamline study information extraction using \method. \textbf{b}, Data extraction accuracy within each field type across four topics. \textbf{c}, Confusion matrix showing the hallucination and missing rates in the data extraction results. \textbf{d}, Result extraction accuracy across topics. \textbf{e}, Result extraction accuracy across clinical endpoints. \textbf{f}, Error analysis of the result extraction. \textbf{g}, Streamline result extraction using \method.}
    \label{fig:exp_extraction_combined}
    \vspace{-1em}
\end{figure}

\subsection*{\method scales data and result extraction from unstructured documents}
\method leverages LLMs to streamline extracting study characteristics such as target therapies, study arm design, and participants' baseline information from involved studies. Specifically, \method refers to the field names and the descriptions from users and use the full content of the study documents in PDF or XML formats as inputs (Fig.~\hyperref[fig:exp_extraction_combined]{4a}). When the free full content is unavailable, \method accepts the user-uploaded content as the input. We developed an evaluation dataset by converting the study characteristic tables from each review paper into data points.  Our dataset comprises 1,334 target data points, including 696 on study design, 353 on population features, and 285 on results. We assessed the data extraction performance using the Accuracy metric. 

\method demonstrated strong extraction performance across various topics (Fig.~\hyperref[fig:exp_extraction_combined]{4b}): it achieved an accuracy of $\acc=0.78$ (95\% confidence interval (CI) = 0.75–0.81) in the Immunotherapy topic, $\acc=0.77$ (95\% CI = 0.72-0.82) in the Radiation/Chemotherapy topic, $\acc=0.72$ (95\% CI = 0.63-0.80) in the Hormone Therapy topic, and $\acc=0.83$ (95\% CI = 0.74-0.90) in the Hyperthermia topic. These results indicate that \method can provide a solid initial data extraction, which human experts can refine. Importantly, each output can be cross-checked by the linked original sources, facilitating verification and further investigation.

Diving deeper into the accuracy across different types of fields, we observed varying performance levels. It performed best in extracting study design information, followed by population details, and showed the lowest accuracy in extracting results (Fig.~\hyperref[fig:exp_extraction_combined]{4b}). For example, in the Immunotherapy topic, \method achieved an accuracy of $\acc=0.95$ (95\% CI = 0.92-0.96) for study design, $\acc=0.74$ (95\% CI = 0.67-0.80) for population data, and $\acc=0.42$ (95\% CI = 0.36-0.49) for results. This variance can be attributed to the prevalence of numerical data in the fields: fields with more numerical data are typically harder to extract accurately. Study design is mostly described in textual format and is directly presented in the documents, whereas population and results often include numerical data such as the number of patients or gender ratios. Results extraction is particularly challenging, often requiring reasoning and transformation to capture values accurately. Given these complexities, it is advisable to scrutinize the extracted numerical data more carefully.

We also evaluated the robustness of \method against hallucinations and missing information  (Fig.~\hyperref[fig:exp_extraction_combined]{4c}). We constructed a confusion matrix detailing instances of hallucinations: false positives (FP) where \method generated data not present in the input document and false negatives (FN) where it failed to extract available target field information. We observed that \method achieved a precision of $\precision=0.994$ for study design, $\precision=0.966$ for population, and $\precision=0.862$ for study results. Missing information was slightly more common than hallucinations, with \method achieving recall rates of $\recall=0.946$ for study design, $\recall=0.889$ for population, and $\recall=0.930$ for study results. The incidence of both hallucinations and missing information was generally low. However, hallucinations were notably more frequent in study results; this often occurred because LLMs could confuse definitions of clinical outcomes, for example, mistaking `overall response' for `complete response.' Nevertheless, such hallucinations are typically manageable, as human experts can easily identify and correct them while reviewing the referenced material.

The challenges in extracting study results primarily stem from (1) identifying the locations that describe the desired outcomes from lengthy papers, (2) accurately extracting relevant numerical values such as patient numbers, event counts, durations, and ratios from the appropriate patient groups, and (3) performing the correct calculations to standardize these values for meta-analysis. In response to these complexities, we developed a specialized pipeline for result extraction (Fig.~\hyperref[fig:exp_extraction_combined]{4g}), where users provide the interested outcome and the cohort definition. \method offers a transparent extraction workflow, documenting the sources of results along with the intermediate reasoning and calculations.

\begin{figure}[htbp]
    \centering
    \includegraphics[width=0.9\linewidth]{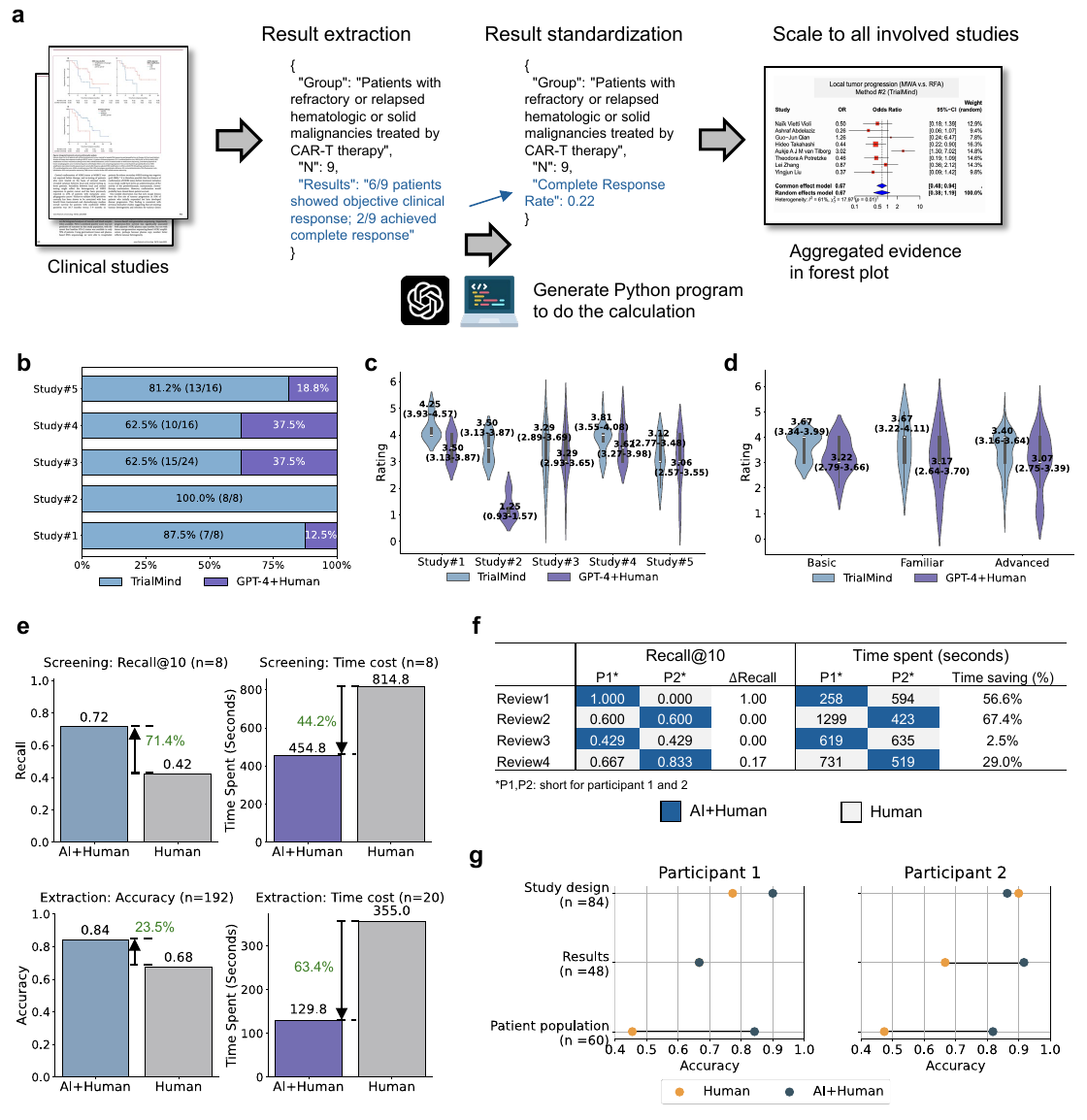}
    \caption{\textbf{Results of human evaluation and user study}. \textbf{a}, \method's result extraction process. \textbf{b}, Winning rate of \method against the GPT-4+Human baseline across studies. \textbf{c}, Violin plots of the ratings across studies. Each plot is tagged with the mean ratings (95\% CI) from all the annotators. \textbf{d},Violin plots of the ratings across annotators with different expertise levels. Each plot is tagged with the mean ratings (95\% CI) from all the studies. \textbf{e}, Overall performance and time cost of study screening and data extraction tasks, respectively. \textbf{f}, Screening time cost and performance across reviews and participants. \textbf{g}, Data extraction accuracy across participants and different types of data.}
    \label{fig:exp_human_eval}
\end{figure}

We compared \method against two generalist LLM baselines, GPT-4 and Sonnet, which were prompted to extract the target outcomes from the full content of the study documents. Since the baselines can only make text extractions, we manually convert them into numbers suitable for meta-analysis~\cite{deeks2010statistical}. This made very strong baselines since they combined LLM extraction with human post-processing. We assessed the performance using the Accuracy metric.

The evaluation conducted across four topics demonstrated the superiority of \method (Fig.~\hyperref[fig:exp_extraction_combined]{4d}). Specifically, in Immunotherapy, \method achieved an accuracy of $\acc=0.70$ (95\% CI 0.62-0.77), while GPT-4 scored $\acc=0.54$ (95\% CI 0.45-0.62). In Radiation/Chemotherapy, \method reached $\acc=0.65$ (95\% CI 0.51-0.76), compared to GPT-4's $\acc=0.52$ (95\% CI 0.39-0.65). For Hormone Therapy, \method achieved $\acc=0.80$ (95\% CI 0.58-0.92), outperforming GPT-4, which scored $\acc=0.50$ (95\% CI 0.30-0.70). In Hyperthermia, \method obtained an accuracy of $\acc=0.84$ (95\% CI 0.71-0.92), significantly higher than GPT-4's $\acc=0.52$ (95\% CI 0.39-0.65). The breakdowns of evaluation results by the most frequent types of clinical outcomes (Fig.~\hyperref[fig:exp_extraction_combined]{4e}) showed \method got fold changes in accuracy ranging from 1.05 to 2.83 and a median of 1.50 over the best baselines. This enhanced effectiveness is largely attributable to \method's ability to accurately identify the correct data locations and apply logical reasoning, while the baselines often produced erroneous initial extractions. 

We analyzed the error cases in our result extraction experiments and identified four primary error types (Fig.~\hyperref[fig:exp_extraction_combined]{4f}). The most common error was `Inaccurate' extraction (n=36), followed by `Extraction failure' (n=27), `Unavailable data' (n=10), and `Hallucinations' (n=3). `Inaccurate' extractions often occurred due to multiple sections ambiguously describing the same field. For example, a clinical study might report the total number of participants receiving CAR-T therapy early in the document and later provide outcomes for a subset with non-small cell lung cancer (NSCLC). The specific results for NSCLC patients are crucial for reviews focused on this subgroup, yet the presence of general data can lead to confusion and inaccuracies in extraction. `Extraction failure' and `Unavailable data' both illustrate scenarios where \method could not retrieve the information. The latter case particularly showcases \method's robustness against hallucinations, as it failed to extract data outside the study's main content, such as in appendices, which were not included in the inputs. Furthermore, errors caused by hallucinations were minor. The outputs were easy to identify and correct through manual inspection since no references were provided.

\subsection*{\method facilitates clinical evidence synthesis via human-AI collaboration}
We selected five systematic review studies as benchmarks and referenced the clinical evidence reported in the target studies. The baseline used GPT-4 with a simple prompting to extract the relevant text pieces that report the target outcome of interest (Methods). Manual calculations were necessary to standardize the data for meta-analysis. In contrast, \method automated the extraction and standardization (Fig.~\hyperref[fig:exp_human_eval]{5a} by (1) extracting the raw result description from the input document and (2) standardizing the results by generating a Python program to assist the calculation. The standardized results from all involved studies are then fed into the R program by human experts to make the aggregated evidence in a forest plot.

We engaged with human annotators to assess the quality of synthesized clinical evidence presented in forest plots. Each annotator was asked to evaluate the evidence quality by comparing it against the evidence reported in the target review and deciding which method, \method or the baseline, produced superior results (Extended Fig.~\ref{fig:human-eval-demo}). Additionally, they rated the quality of the synthesized clinical evidence on a scale of 1 to 5. The assignment of our method and the baseline was randomized to ensure objectivity. The results highlighted \method's superior performance compared to the direct use of GPT-4 for clinical evidence synthesis (Fig.~\hyperref[fig:exp_human_eval]{5b}). We calculated the winning rate of \method versus the baseline across the five studies. The results indicate a consistent preference by annotators for the evidence synthesized by \method over that of the baseline. Specifically, \method achieved winning rates of 87.5\%, 100\%, 62.5\%, 62.5\%, and 81.2\%, respectively. The baseline's primary shortcoming stemmed from the initial extraction step, where GPT-4 often failed to identify the relevant sources without well-crafted prompting. Therefore, the subsequent manual post-processing was unable to rectify these initial errors.

In addition, we illustrated the ratings of \method and the baseline across studies (Fig.~\hyperref[fig:exp_human_eval]{5c}). We found \method was competent as the GPT-4+Human baseline and outperformed the baseline in many scenarios. For example, \method obtained the mean rating of 4.25 (95\% CI 3.93-4.57) in Study \#1 while the baseline obtained 3.50 (95\% CI 3.13-3.87). In Study \#2, \method yielded 3.50 (95\% CI 3.13-3.87) while the baseline yielded 1.25 (95\% CI 0.93-1.57). The performance of the two methods was comparable in the remaining three studies. These results highlight \method as a highly effective alternative to conventional LLM usage in evidence synthesis, streamlining data extraction and processing while maintaining the critical benefit of human oversight.

We requested that annotators self-assess their expertise level in clinical studies, classifying themselves into three categories: `Basic', `Familiar', and `Advanced'. The typical profile ranges from computer scientists at the basic level to medical doctors at the advanced level. We then analyzed the ratings given to both methods across these varying expertise levels (Fig.~\hyperref[fig:exp_human_eval]{5d}). We consistently observed higher ratings for \method than the baseline across all groups. Annotators with basic knowledge tended to provide more conservative ratings, while those with more advanced expertise offered a wider range of evaluations. For instance, the `Basic' group provided average ratings of 3.67 (95\% CI 3.34-3.39) for \method compared to 3.22 (95\% CI 2.79-3.66) for the baseline. The `Advanced' group rated \method at an average of 3.40 (95\% CI 3.16-3.64) and the baseline at 3.07 (95\% CI 2.75-3.39).

We conducted user studies to compare the quality and time efficiency between purely manual efforts and human-AI collaboration using \method. Two participants were involved in both study screening and data extraction tasks. For the screening task, each participant was assigned 4 systematic review papers, with 100 candidate citations identified for each review. The participants were asked to select the 10 most likely relevant citations from the candidate pool. Each participant was provided with 2 candidate sets pre-ranked by \method and 2 unranked sets. The participants also recorded the time taken to complete the screening process for each set. For the data extraction task, each participant was given 10 clinical studies. They manually extracted the target information for 5 of these studies. For the other 5, \method was first used to perform an initial extraction, and the participants were required to verify and correct the extracted results. The time taken for the extraction process was reported for each study.

In Fig.~\hyperref[fig:exp_human_eval]{5e}, we present the average performance and time cost for the AI+Human and Human-only approaches across both the study screening and data extraction tasks. The results demonstrate that the AI+Human approach consistently outperforms the Human-only approach. For the screening tasks, AI+Human achieved a 71.4\% relative improvement in Recall, while reducing time by 44.2\% compared to the Human-only arm. This underscores the significant advantage of \method in accelerating the study screening process while also improving its quality. Similarly, for the data extraction tasks, the AI+Human approach improved extraction accuracy by 23.5\% on average, with a 63.4\% reduction in time required.

Detailed results of screening time and performance are shown in Fig.~\hyperref[fig:exp_human_eval]{5f}, where two reviews showed the AI+Human approach achieving the same Recall as the Human-only arm with notable time savings, and in two other reviews, AI+Human achieved higher Recall with less time. From Fig.~\hyperref[fig:exp_human_eval]{5g}, we see that the AI+Human approach delivered better or comparable accuracy across all three types of data, with the smallest gap in ``Study design". This is likely because study design information is often readily available in the study abstract, making it relatively easier for humans to extract. In contrast, the other two data types are embedded deeper within the main content, which can sometimes make it challenging for human readers to locate the correct information.

\section*{Discussion}\label{sec:discussion}
Clinical evidence forms the bedrock of evidence-based medicine, crucial for enhancing healthcare decisions and guiding the discovery and development of new therapies. It often comes from a systematic review of diverse studies found in the literature, encompassing clinical trials and retrospective analyses of real-world data. Yet, the burgeoning expansion of literature databases presents formidable challenges in efficiently identifying, summarizing, and maintaining the currency of this evidence. For instance, a study by the US Agency for Healthcare Research and Quality (AHRQ) found that half of 17 clinical guidelines became outdated within a couple of years~\cite{shekelle2001validity}.

The rapid development of large language models (LLMs) and AI technologies has generated considerable interest in their potential applications in clinical research~\cite{hutson2024ai,wang2023pytrial}. However, most of them focused on an individual aspect of the clinical evidence synthesis process, such as literature search~\cite{jin2024pubmed,scells2017test}, citation screening~\cite{wallace2010semi,kanoulas2018clef,trikalinos2019large}, quality assessment~\cite{vsuster2023automating}, or data extraction~\cite{yun2024automatically,schmidt2021data}. In addition, implementing these models in a manner that is collaborative, transparent, and trustworthy poses significant challenges, especially in critical areas such as medicine~\cite{zhang2024leveraging}. For instance, when utilizing LLMs to summarize evidence from multiple studies, the descriptive summaries often usually merely echo the findings verbatim, omit crucial details, and fail to adhere to established best practices~\cite{shaib2023summarizing}. Besides, when given a set of studies that are irrelevant to the research question, LLMs are prone to produce hallucinations and hence cause misleading evidence~\cite{joseph2024factpico}. This challenge highlights the need for an integrated pipeline, involving the study search and screening stages, to strategically pick the target studies for analysis~\cite{ramprasad2023automatically,chelli2024hallucination}, or enhanced with human-AI collaboration~\cite{spillias2023human}.

This study introduces a clinical evidence synthesis pipeline enhanced by LLMs, named \method. This pipeline is structured in accordance with established medical systematic review protocols, involving steps such as study searching, screening, data/result extraction, and evidence synthesis. At each stage, human experts have the capability to access, monitor, and modify intermediate outputs. This human oversight helps to eliminate errors and prevents their propagation through subsequent stages. Unlike approaches that solely depend on the knowledge of LLMs, \method integrates human expertise through in-context learning and chain-of-thought prompting. Additionally, \method extends external knowledge sources to its outputs through retrieval-augmented generation and leveraging external computational tools to enhance the LLM's reasoning and analytical capabilities. Comparative evaluations of \method and traditional LLM approaches have demonstrated the advantages of this system design in LLM-driven applications within the medical field.

This study also has several limitations. First, despite incorporating multiple techniques, LLMs may still make errors at any stage. Therefore, human oversight and verification remain crucial when implementing \method in practical settings. Second, the prompts used in \method were developed based on prompt engineering experience, suggesting potential for performance enhancement through advanced prompt optimization or by fine-tuning the underlying LLMs to suit specific tasks better. Third, while \method demonstrated effectiveness in study search, screening, and data extraction, the dataset used was limited in size due to the high costs associated with human labeling. Future research could expand on these findings with larger datasets to further validate the method's effectiveness. Fourth, the study coverage was restricted to publicly available sources from PubMed Central, which provides structured PDFs and XMLs. Many relevant studies are either not available on PubMed or are in formats that entail OCR algorithms as preprocessing, indicating a need for further engineering to incorporate broader data sources. Fifth, although \method illustrated the potential of using advanced LLMs like GPT-4 to streamline clinical evidence synthesis, developing techniques to adapt the pipeline for use with other LLMs could increase its applicability. Finally, while the use of LLMs like GPT-4 can accelerate study screening and data extraction, the associated costs and processing times may present bottlenecks in some scenarios. Future enhancements that improve efficiency or utilize localized, specialized smaller models could increase practical utility.

LLMs have made significant strides in AI applications. \method exemplifies a crucial aspect of system engineering in LLM-driven pipelines, facilitating the practical, robust, and transparent use of LLMs. We anticipate that \method will benefit the medical AI community by fostering the development of LLM-driven medical applications and emphasizing the importance of human-AI collaboration.

\section*{Methods}\label{sec:methods}

\subsection*{Description of the \dataset Dataset}
The overall flowchart for the study identification and screening process in building \dataset is illustrated in Extended Fig.~\ref{fig:metasyns_screening}.

\paragraph{Database search and initial filtering}
We undertook a comprehensive search on the PubMed database for meta-analysis papers related to cancer. The Boolean search terms were specifically chosen to encompass a broad spectrum of cancer-related topics. These terms included ``cancer", ``oncology", ``neoplasm", ``carcinoma", ``melanoma", ``leukemia", ``lymphoma", and ``sarcoma". Additionally, we incorporated terms related to various treatment modalities such as ``therapy", ``treatment", ``chemotherapy", ``radiation therapy", ``immunotherapy", ``targeted therapy", ``surgical treatment", and ``hormone therapy". To ensure that our search was exhaustive yet precise, we also included terms like ``meta-analysis" and "systematic review" in our search criteria.

This initial search yielded an extensive pool of 46,192 results, reflecting the vast research conducted in these areas. We applied specific filters to refine these results and ensure relevance and quality. We focused on articles where PMC Full text was available and specifically categorized under ``Meta-Analysis". Further refinement was done by restricting the time frame of publications to those between January 1, 2020, and January 1, 2023. We also narrowed our focus to studies conducted on humans and those available in English. This filtration process was critical in distilling the initial results into a more manageable and focused collection of 2,691 papers. 

\paragraph{Refinement}
Building upon our initial search, we employed further refinement techniques using both MeSH terms and specific keywords. The MeSH terms were carefully selected to target papers precisely relevant to various forms of cancer. These terms included ``cancer", ``tumor", ``neoplasms", ``carcinoma", ``myeloma", and ``leukemia". This focused approach using MeSH terms effectively reduced our selection to 1,967 papers.

To further dive in on papers investigating cancer therapies, we utilized many keywords derived from the National Cancer Institute’s ``Types of Cancer Treatment" list. This approach was multi-faceted, with each set of keywords targeting a specific category of cancer therapy. For chemotherapy, we included terms like ``chemotherapy", ``chemo", and related variations. In the realm of hormone therapy, we searched for phrases such as "hormone therapy", "hormonal therapy", and similar terms. The keyword group for hyperthermia encompassed terms like ``hyperthermia", ``microwave", ``radiofrequency", and related technologies. For cancer vaccines, we included keywords such as ``cancer vaccines", ``cancer vaccine", and other related terms. The search for immune checkpoint inhibitors and immune system modulators was comprehensive, including terms like ``immune checkpoint inhibitors", ``immunomodulators", and various cytokines and growth factors. Lastly, our search for monoclonal antibodies and T-cell transfer therapy included relevant terms like ``monoclonal antibodies", ``t-cell therapy", ``car-t", and other related phrases.

The careful application of keyword filtering played a crucial role in narrowing down our pool of research papers to a more focused and relevant set of 352. It represents a diverse and meaningful collection of studies in cancer therapy, highlighting a range of innovative and impactful research within this field.

\paragraph{Manual screening of titles and abstracts}
Then, we manually screened titles and abstracts, applying a rigorous classification and sorting methodology. The remaining papers were first categorized based on the type of cancer treatment they explored. We then organized these papers by their citation count to gauge their impact and relevance in the field. Our selection criteria aimed to enhance the quality and relevance of our final dataset. We prioritized papers that focused on the study of treatment effects, such as safety and efficacy, of various cancer interventions. We preferred studies that compared individual treatments against a control group, as opposed to those examining the effects of combined therapies (e.g., Therapy A+B vs. A only). To build a list of representative meta-analyses, we needed to ensure diversity in the target conditions under each treatment category.

Further, we favored studies that involved a larger number of individual studies, providing a broader base of evidence. However, we excluded network analysis studies and meta-analyses that focused solely on prognostic and predictive effects, as they did not align with our primary research focus. To maintain a balanced representation, we limited our selection to a maximum of three papers per treatment category. This process culminated in a final dataset comprising 100 systematic review papers. This curated collection forms the backbone of our analysis, ensuring a concentrated and pertinent selection of high-quality studies directly relevant to our research objectives.

\subsection*{LLM Prompting}
Prompting steers LLMs to conduct the target task without training the underlying LLMs. \method proceeds clinical evidence synthesis in multiple steps associated with a series of prompting techniques.

\paragraph{In-context learning} LLMs exhibit a profound ability to comprehend input requests and adhere to provided instructions during generation. The fundamental concept of in-context learning (ICL) is to enable LLMs to learn from examples and task instructions within a given context at inference time~\cite{brown2020language}. Formally, for a specific task, we define $T$ as the task prompt, which includes the task definition, input format, and desired output format. During a single inference session with input $X$, the LLM is prompted with $P(T, X)$, where $P(\cdot)$ is a transformation function that restructures the task definition $T$ and input $X$ into the prompt format. The output $\hat{X}$ is then generated as $\hat{X} = \llm(P(T, X))$.

\paragraph{Retrieval-augmented generation} LLMs that rely solely on their internal knowledge often produce erroneous outputs, primarily due to outdated information and hallucinations. This issue can be mitigated through retrieval-augmented generation (RAG), which enhances LLMs by dynamically incorporating external knowledge into their prompts during generation~\cite{lewis2020retrieval}. We denote $R_K(\cdot)$ as the retriever that utilizes the input $X$ to source relevant contextual information through semantic search. $R_K(\cdot)$ enables the dynamic infusion of tailored knowledge into LLMs at inference time.

\paragraph{Chain-of-thought}  Chain-of-though (CoT) guides LLMs in solving a target task in a step-by-step manner in one inference, hence handling complex or ambiguous tasks better and inducing more accurate outputs~\cite{wei2022chain}. CoT employs the function $P_{\text{CoT}}(\cdot)$ to structure the task $T$ into a series of chain-of-thought steps $\{S_1, S_2, \dots, S_T\}$. As a result, we obtain $\{\hat{X}_S^1, \dots, \hat{X}_S^T\} = \text{LLM}(P_{\text{CoT}}(T,X))$, all produced in a single inference session. This is rather critical when we aim to elicit the thinking process of LLM and urge it in self-reflection to improve its response. For instance, we may ask LLM to draft the initial response in the first step and refine it in the second.

\paragraph{LLM-driven pipeline} Clinical evidence synthesis involves a multi-step workflow as outlined in the PRISMA statement~\cite{page2021prisma}. It can be generally outlined as identifying and screening studies from databases, extracting characteristics and results from individual studies, and synthesizing the evidence. To enhance each step's performance, task-specific prompts can be designed for an LLM to create an LLM-based module. This results in a chain of prompts that effectively addresses a complex problem, which we call LLM-driven workflow. Specifically, this approach breaks down the entire meta-analysis process into a sequence of $N $ tasks, denoted as $\mathcal{T} = \{T_1,\dots,T_N\} $. In the workflow, the output from one task, $\hat{X}_n $, serves as the input for the next, $\hat{X}_{n+1} = \text{LLM}(P(T_n, \hat{X}_n)) $. This modular decomposition improves LLM performance by dividing the workflow into more manageable segments, increases transparency, and facilitates user interaction at various stages.

Incorporating these techniques, the formulation of \method for any subtask can be represented as:
\begin{equation}
    \hat{X}_{n+1} = \text{LLM}(P(T_{n}, X_{n}), R_K(X_{n})), \ \forall n = 1, \dots, N,
\end{equation}
where $R_K(\cdot)$ are optional.

\subsection*{Implementation of \method}
All experiments were run in Python v.3.9. Detailed software versions are: pandas v2.2.2; numpy v1.26.4; scipy v1.13.0; scikit-learn v1.4.1.post1; openai v1.23.6; langchain v0.1.16; boto3 v1.34.94; pypdf v4.2.0; lxml v5.2.1 and chromadb v0.5.0 with Python v.3.9.

\paragraph{LLMs} We included GPT-4 and Sonnet in our experiments. GPT-4~\cite{openai2024gpt4} is regarded as a state-of-the-art LLM and has demonstrated strong performances in many natural language processing tasks (version: gpt-4-0125-preview). Sonnet~\cite{anthropic2023claude} is an LLM developed by Anthropic, representing a more lightweight but also very capable LLM (version: anthropic.claude-3-sonnet-20240229-v1:0 on AWS Bedrock). Both models support long context lengths (128K and 200K), enabling them to process the full content of a typical PubMed paper in a single inference session.

\paragraph{Research question inputs} \method processes research question inputs using the PICO (Population, Intervention, Comparison, Outcome) framework to define the study's research question. In our experiments, the title of the target review paper served as the general description. Subsequently, we extracted the PICO elements from the paper's abstract to detail the specific aspects of the research question.

\paragraph{Literature search} \method is tailored to adhere to the established guidelines~\cite{page2021prisma} in conducting literature search and screening for clinical evidence synthesis. In the literature search stage, the key is formulating Boolean queries to retrieve a comprehensive set of candidate studies from databases. These queries, in general, are a combination of treatment, medication, and outcome terms, which can be generated by LLM using in-context learning. However, direct prompting can yield low recall queries due to the narrow range of user inputs and the LLMs' tendency to produce incorrect queries, such as generating erroneous MeSH (Medical Subject Headings) terms~\cite{wang2023can}. To address these limitations, \method incorporates RAG to enrich the context with knowledge sourced from PubMed, and employs CoT processing to facilitate a more exhaustive generation of relevant terms.

Specifically, the literature search component has two main steps: initial query generation and then query refinement. In the first step, \method prompts LLM to create the initial boolean queries derived from the input PICO to retrieve a group of studies (Prompt in Extended Fig.~\ref{fig:prompt_initial_query_generation}). The abstracts of these studies then enrich the context for refining the initial queries, working as RAG. In addition, we used CoT to enhance the refinement by urging LLMs to conduct multi-step reasoning for self-reflection enhancement (Prompt in Extended Fig.~\ref{fig:prompt_query_expansion_refinement}). This process can be described as 
\begin{equation}
    \{\hat{X}_{S}^{1},  \hat{X}_{S}^{2},  \hat{X}_{S}^{3}\} = \text{LLM}(P_{\text{CoT}}(T_{\text{LS}},X,R_K(X))),
\end{equation}
where $X$ denotes the input PICO; $R_K(X)$ is the set of abstracts of the found studies; $T_{\text{LS}}$ is the definition of the query generation task for literature search. For the output, the first sub-step $\hat{X}_S^1$ indicates a complete set of terms identified in the found studies; the second $\hat{X}_S^2$ indicates the subset of $\hat{X}_S^1$ by filtering out the irrelevant; and the third $\hat{X}_S^3$ indicates the extension of $\hat{X}_S^2$ by self-reflection and adding more augmentations. In this process, LLM will produce the outputs for all three substeps in one pass, and \method takes $\hat{X}_S^3$ as the final queries to fetch the candidate studies.

\paragraph{Study screening} \method follows PRISMA to take a transparent approach for study screening. It creates a set of eligibility criteria based on the input PICO as the basis for study selection (Prompt in Extended Fig.~\ref{fig:prompt_study_eligibility_generation}), produced by
\begin{equation}
    \hat{X}_{\text{EC}} = \text{LLM}(P(T_{\text{EC}}, X)),
\end{equation} 
where $\hat{X}_{\text{EC}} = \{E_1, E_2,\dots, E_M\}$ is the $M$ generated eligibility criteria; $X$ is the input PICO; and $T_{\text{EC}}$ is the task definition of criteria generation. Users are given the opportunity to modify these generated criteria, further adjusting to their needs.

Based on $\hat{X}_{\text{EC}}$, \method embarks the parallel processing for the candidate studies. For $i$-th study $F_i$, the eligibility prediction is made by LLM as (Prompt in Extended Fig.~\ref{fig:prompt_eligibility_assessment})
\begin{equation}
    \{I_i^1,\dots,I_i^M\} = \text{LLM}(P(F_i, X, T_{\text{SC}},\hat{X}_{\text{EC}})),
\end{equation}
where $T_{\text{SC}}$ is the task definition of study screening; $F_i$ is the study $i$'s content; $I_i^m \in \{-1,0,1\}, \forall m=1,\dots,M$ is the prediction of study $i$'s eligibility to the $m$-th criterion. Here, $-1$ and $1$ mean ineligible and eligible, $0$ means uncertain, respectively. These predictions offer a convenient way for users to inspect the eligibility and select the target studies by altering the aggregation strategies. $I_i^m$ can be aggregated to offer an overall relevance of each study, such as $\hat{I}_i = \sum_m I_i^m$. Users are also encouraged to extend the criteria set or block the predictions of some criteria to make customized rankings during the screening phase.

\paragraph{Data extraction} Study data extraction is an open information extraction task that requires the model to extract specific information based on user inputs and handle long inputs, such as the full content of a paper. LLMs are particularly well-suited for this task because (1) they can perform zero-shot learning via in-context learning, eliminating the need for labeled training data, and (2) the most advanced LLMs can process extremely long inputs.   As such, the \method framework is engineered to streamline data extraction from structured or unstructured study documents using LLMs. 

For the specified data fields to be extracted, \method prompts LLMs to locate and extract the relevant information (Prompt in Extended Fig.~\ref{fig:prompt_study_extraction}). These data fields include (1) study characteristics such as study design, sample size, study type, and treatment arms; (2) population baselines; and (3) study findings. In general, the extraction process can be described as
\begin{equation}\label{eq:data_extract}
    \{\hat{X}^1_{\text{EX}}, \dots, \hat{X}^K_{\text{EX}}\} = \text{LLM}(P(F,C,T_{\text{EX}})),
\end{equation}
where $ F $ represents the full content of a study; $ T_{\text{EX}} $ defines the task of data extraction; and $ C = \{C_1, C_2, \dots, C_K\} $ comprises the series of data fields targeted for extraction. $C_k$ is the user input natural language description of the target field, e.g., ``the number of participants in the study". The input content $ F $ is segmented into distinct chunks, each marked by a unique identifier. The outputs, denoted as $ \hat{X}^k_{\text{EX}} = \{V^k, B^k\} $, include the extracted values $ V $ and the indices $ B $ that link back to their respective locations in the source content. Hence, it is convenient to check and correct mistakes made in the extraction by sourcing the origin. The extraction can also be easily scaled by making paralleled calls of LLMs.

\paragraph{Result extraction} Our analysis indicates that data extraction generally performs well for study design and population-related fields; however, extracting study results presents challenges. Errors frequently arise due to the diverse presentation of results within studies and subtle discrepancies between the target population and outcomes versus those reported. For instance, the target outcome is the risk ratios (treatment versus control) regarding the incidence of adverse events (AEs), while the study reports AEs among many groups separately. Or, the target outcome is the incidence of severe AEs, which implicitly correspond to those with grade III and more, while the study reports all grade AEs. To overcome these challenges, we have refined our data extraction process to create a specialized result extraction pipeline that improves clinical evidence synthesis. This enhanced pipeline consists of three crucial steps: (1) identifying the relevant content within the study (Prompt in Extended Fig.~\ref{fig:prompt_initial_result_extraction}), (2) extracting and logically processing this content to obtain numerical values (Prompt in Extended Fig.~\ref{fig:prompt_result_formatting}), and (3) converting these values into a standardized tabular format (Prompt in Extended Fig.~\ref{fig:prompt_result_standardization}).

Steps (1) and (2) are conducted in one pass using CoT reasoning as
\begin{equation}
    \{\hat{X}^1_{\text{RE},S},\hat{X}^2_{\text{RE},S}\} = \text{LLM}(P_{\text{CoT}}(X, O, F, T_{\text{RE}})),
\end{equation}
where $O$ is the natural language description of the clinical endpoint of interest and $T_{\text{RE}}$ is the task definition of result extraction. In the outputs, $\hat{X}_{\text{RE},S}^1$ represents the raw content captured from the input content $F$ regarding the clinical outcomes;  $\hat{X}_{\text{RE},S}^2$ represents the elicited numerical values from the raw content, such as the number of patients in the group, the ratio of patients encountering overall response, etc. In step (3), \method writes Python code to make the final calculation to convert $\hat{X}_{\text{RE},S}^2$ to the standard tabular format.
\begin{equation}
    \hat{X}_{\text{RE}} = \texttt{exec}(\text{LLM}(P(X, O, T_{\text{PY}},\hat{X}_{\text{RE},S}^2)), \hat{X}_{\text{RE},S}^2 ).
\end{equation}
In this process, \method adheres to the instructions in $ T_{\text{PY}} $ to generate code for data processing. This code is then executed, using $ \hat{X}_{\text{RE},S}^2 $ as input, to produce the standardized result $ \hat{X}_{\text{RE}} $.  An example code snippet made to do this transformation is shown in Extended Fig.~\ref{fig:demo_codegen}. This approach facilitates verification of the extracted results by allowing for easy backtracking to $ \hat{X}_{\text{RE},S}^1 $. Additionally, it ensures that the calculation process remains transparent, enhancing the reliability and reproducibility of the synthesized evidence.

\subsection*{Experimental setup}

\paragraph{Literature search and screening}
In our literature search experiments, we assessed performance using the overall Recall, aiming to evaluate the effectiveness of different methods in identifying all relevant studies from the PubMed database using APIs~\cite{ncbi2002biotech}. For literature screening, we measured efficacy using Recall@20 and Recall@50, which gauge how well the methods can prioritize target studies at the top of the list, thereby facilitating quicker decisions about which studies to include in evidence synthesis. We constructed the ranking candidate set for each review paper by initially retrieving studies through \method, then refining this list by ranking the relevance of these studies to the target review's PICO elements using OpenAI embeddings. The top 2,000 relevant studies were kept. We then ensured all target papers were included in the candidate set to maintain the integrity of our ground-truth data. The final candidate set was then deduplicated to be ranked by the selected methods.

In the criteria analysis experiment, we utilized Recall@200 to assess the impact of each criterion. This was done by first computing the relevance prediction using all eligibility predictions and then recalculating it without the eligibility prediction for the specific criterion in question. The difference in Recall@200 between these two relevance predictions, denoted as $\Delta \recall$, indicates the criterion's effect. A larger $\Delta \recall$ suggests that the criterion plays a more significant role in influencing the ranking results.

\paragraph{Data extraction and result extraction} To evaluate performance, we measured the accuracy of the values extracted by \method against the groundtruth. We used the study characteristic tables from the review papers as our test set. Each table's column names served as input field descriptions for \method. We manually downloaded the full content for the studies listed in the characteristic table. To verify the accuracy of the extracted values, we enlisted three annotators who manually compared them against the data reported in the original tables.

We also measured the performance of result extraction using accuracy. The annotators were asked to carefully read the extracted results and compare them to the results reported in the original review paper. For the error analysis of \method, the annotators were asked to check the sources to categorize the errors for one of the reasons: inaccurate, extraction failure, unavailable data, or hallucination. We designed a vanilla prompting strategy for GPT-4 and Sonnet models to set the baselines for the result extraction. Specifically, the prompt was kept minimal, as ``Based on the \{paper\}, tell me the \{outcome\} from the input study for the population \{cohort\}", where \{paper\} is the placeholder for the paper's content;  \{outcome\} is the for the target endpoint; \{cohort\} is the for the target population's descriptions, including conditions and characteristics. The responses from these prompts were typically in free text, from which annotators manually extracted result values to evaluate the baselines' performance.

\paragraph{Evidence synthesis} In evidence synthesis, we processed the input data using R and the `meta' package to make the forest plots and the pooled results based on the standardized result values. This is for both \method and the baselines. Nonetheless, for the baseline, the annotators also need to manually extract the result values and standardize the values to make them ready for meta-analysis, which forms the GPT-4+Human baseline in the experiments.

We engaged two groups of annotators for our evaluation: (1) three computer scientists with expertise in AI applications for medicine, and (2) five medical doctors to assess the generated forest plots. Each annotator was asked to evaluate five review studies. For each review, we randomly presented forest plots generated by both the baseline and \method. The annotators were required to determine how closely each generated plot aligned with a reference forest plot taken from the target review paper. Additionally, they were asked to judge which method, the baseline or \method, produced better results in a win/lose assessment. Extended Fig.~\ref{fig:human-eval-demo} demonstrates the user interface for this study, which was created with Google Forms.





\clearpage

\captionsetup[table]{name=Extended Table}
\setcounter{figure}{0}
\renewcommand*{\figurename}{Extended Fig.}

\begin{appendices}

\begin{figure}
    \centering
    \includegraphics[width=\linewidth]{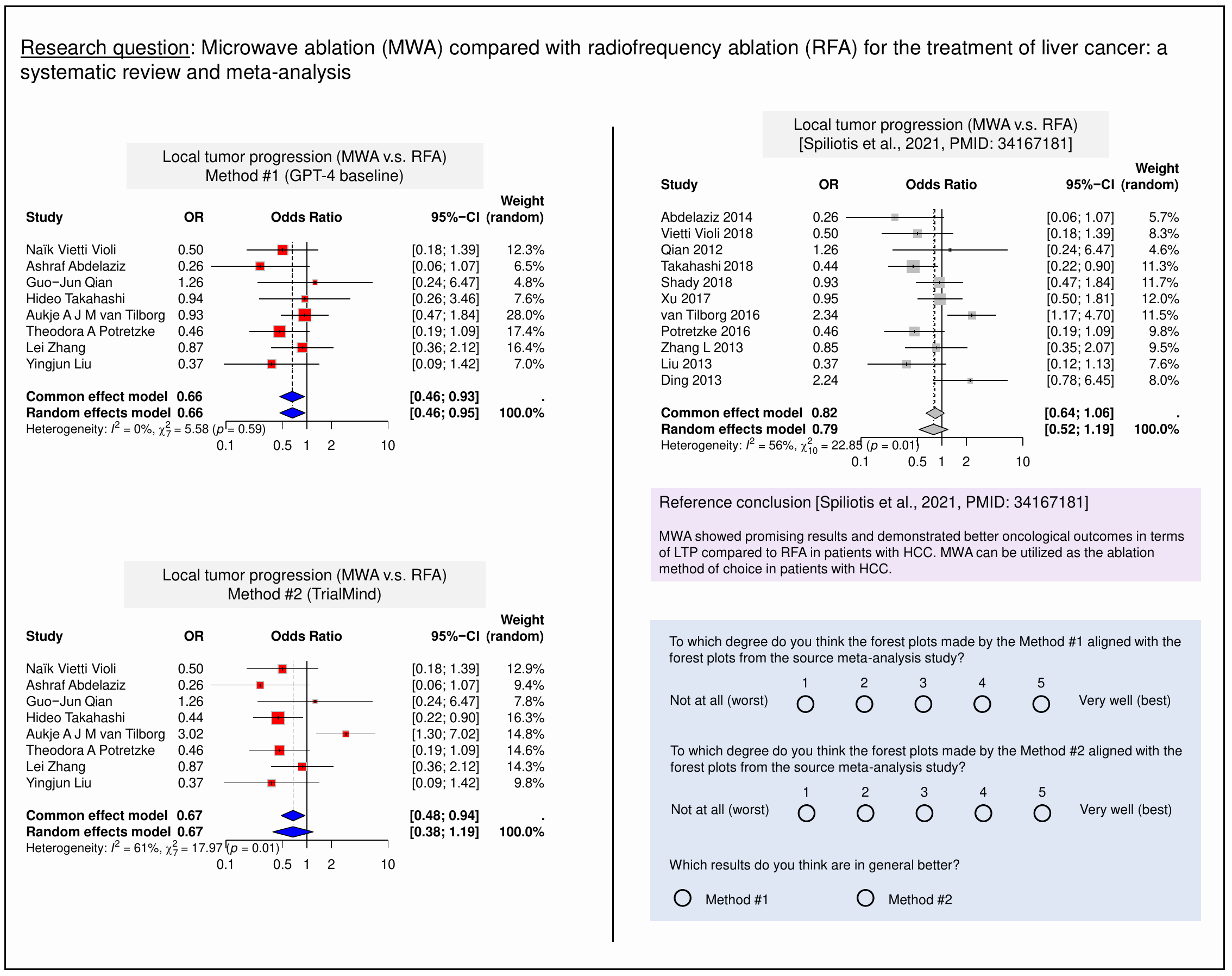}
    \caption{The study design compares the synthesized clinical evidence from the baseline and \method via human evaluation. }
    \label{fig:human-eval-demo}
\end{figure}

\clearpage

\begin{figure}[t]
    \centering
    \includegraphics[width=\linewidth]{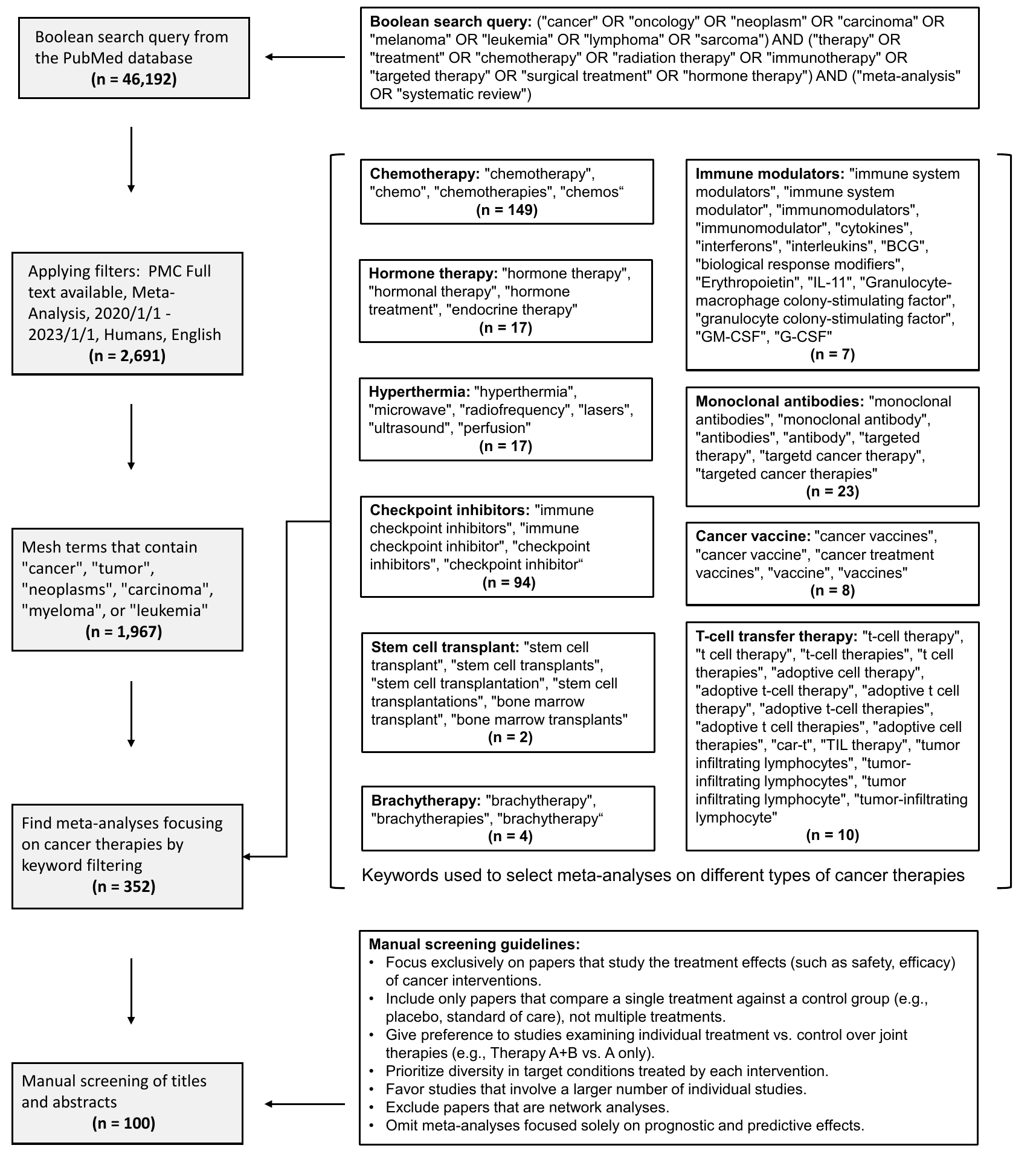}
    \caption{The flowchart of the screening process of meta-analyses involved in the \dataset dataset.}
    \label{fig:metasyns_screening}
\end{figure}

\clearpage

\begin{figure}[t]
    \centering
    \includegraphics[width=\linewidth]{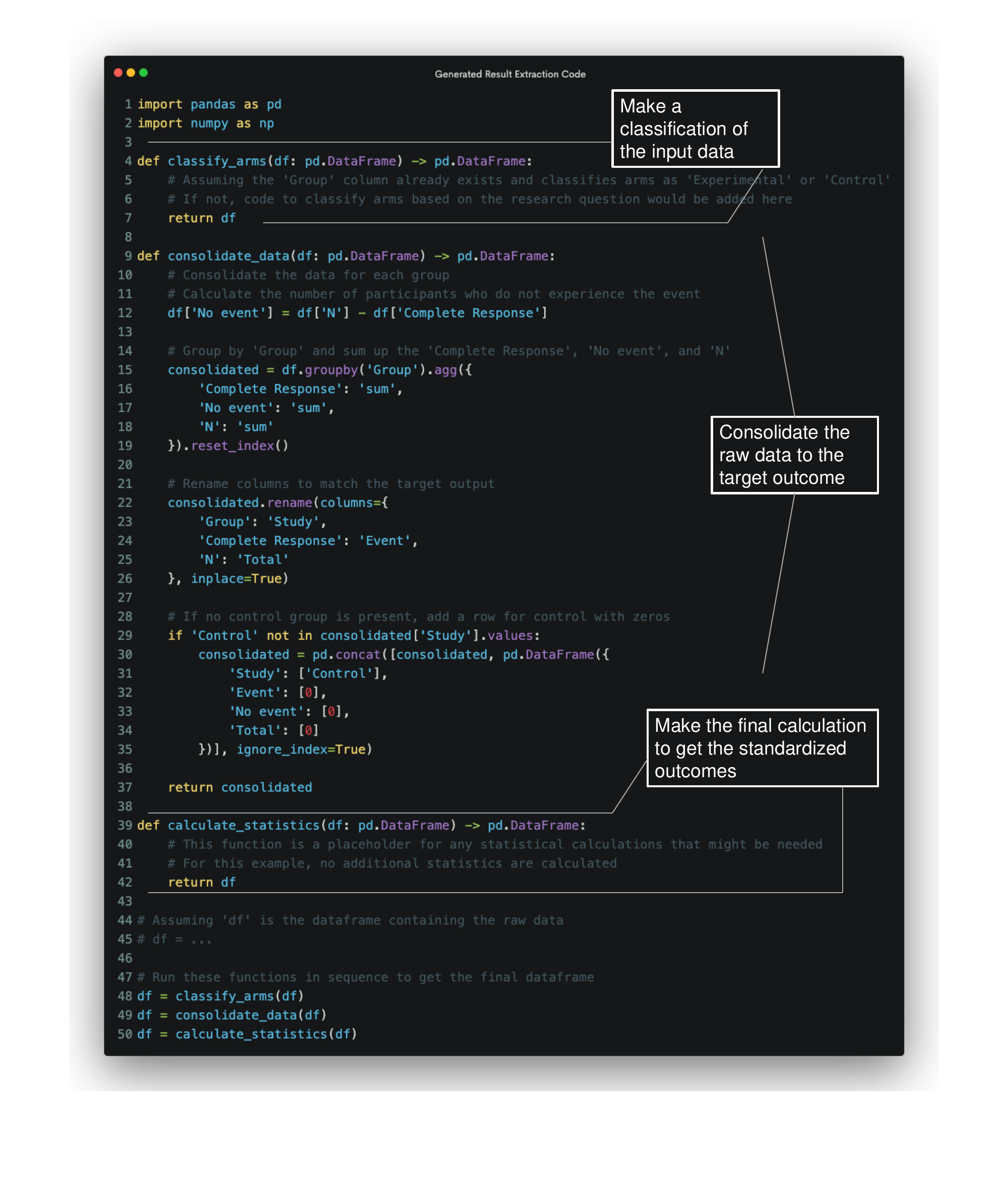}
    \caption{The example Python code made by \method when converting the extracted result values to standardized tabular form.}
    \label{fig:demo_codegen}
\end{figure}

\clearpage

\begin{figure}[!htbp]
    \centering
    \includegraphics[width=1.0\linewidth]{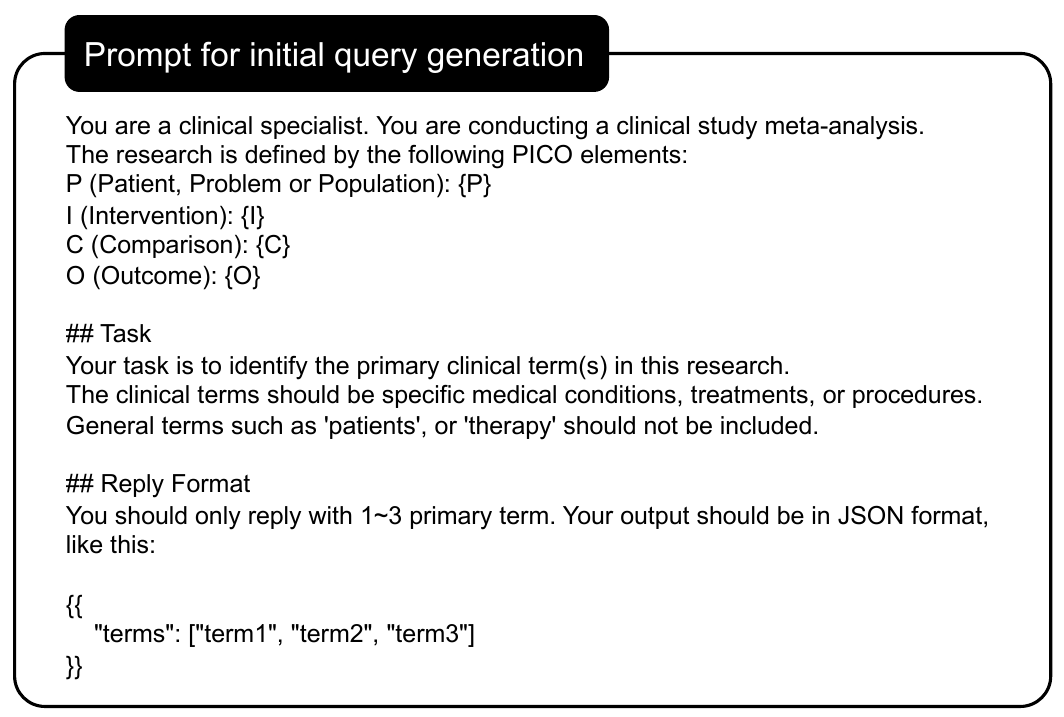} 
    \caption{Prompt for generating initial search queries in the literature search.}
    \label{fig:prompt_initial_query_generation}
\end{figure}

\begin{figure}[!htbp]
    \centering
    \includegraphics[width=0.9\linewidth]{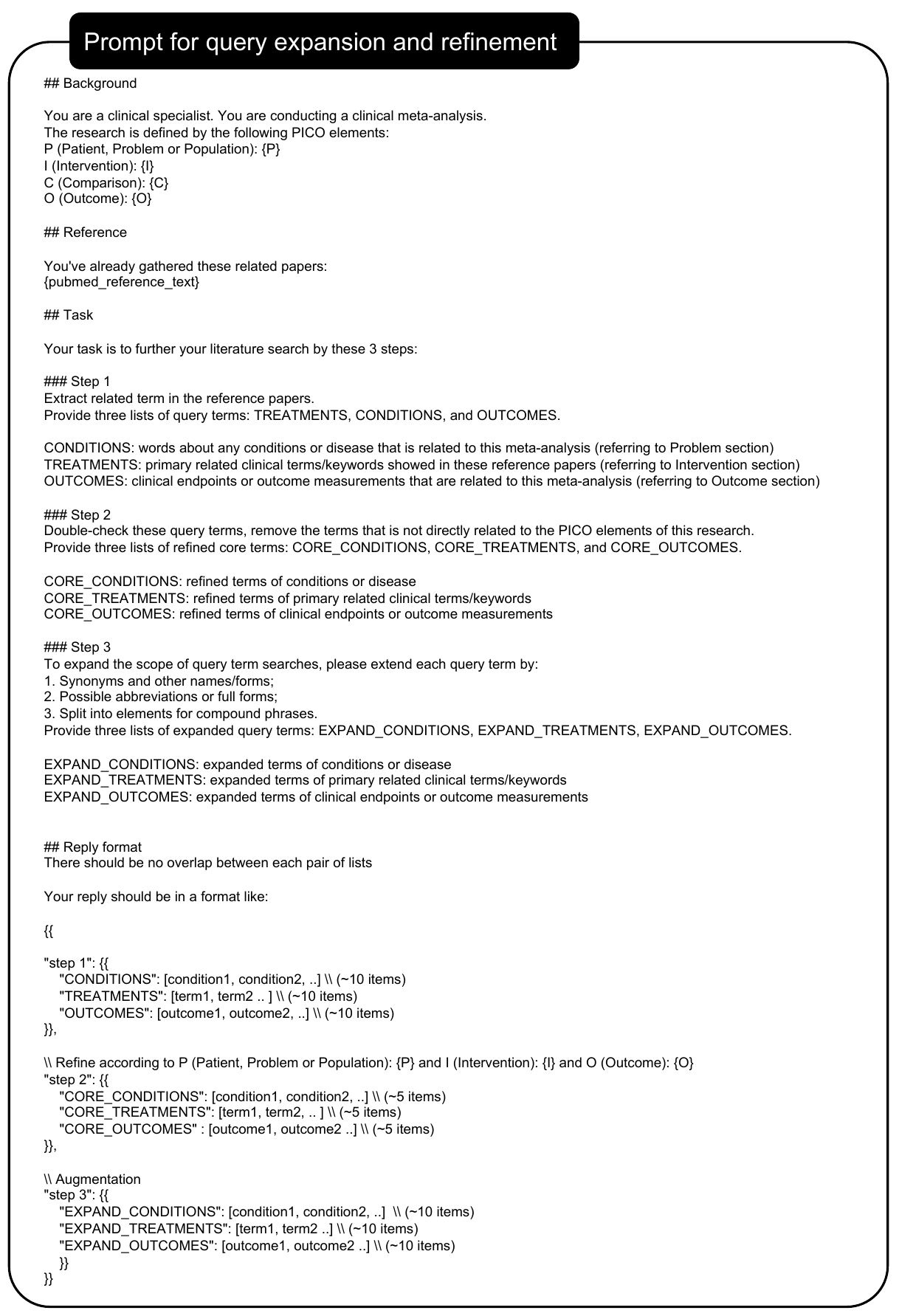} 
    \caption{Prompt for expanding and refining the initial search queries in the literature search.}
    \label{fig:prompt_query_expansion_refinement}
\end{figure}

\begin{figure}[!htbp]
    \centering
    \includegraphics[width=0.9\linewidth]{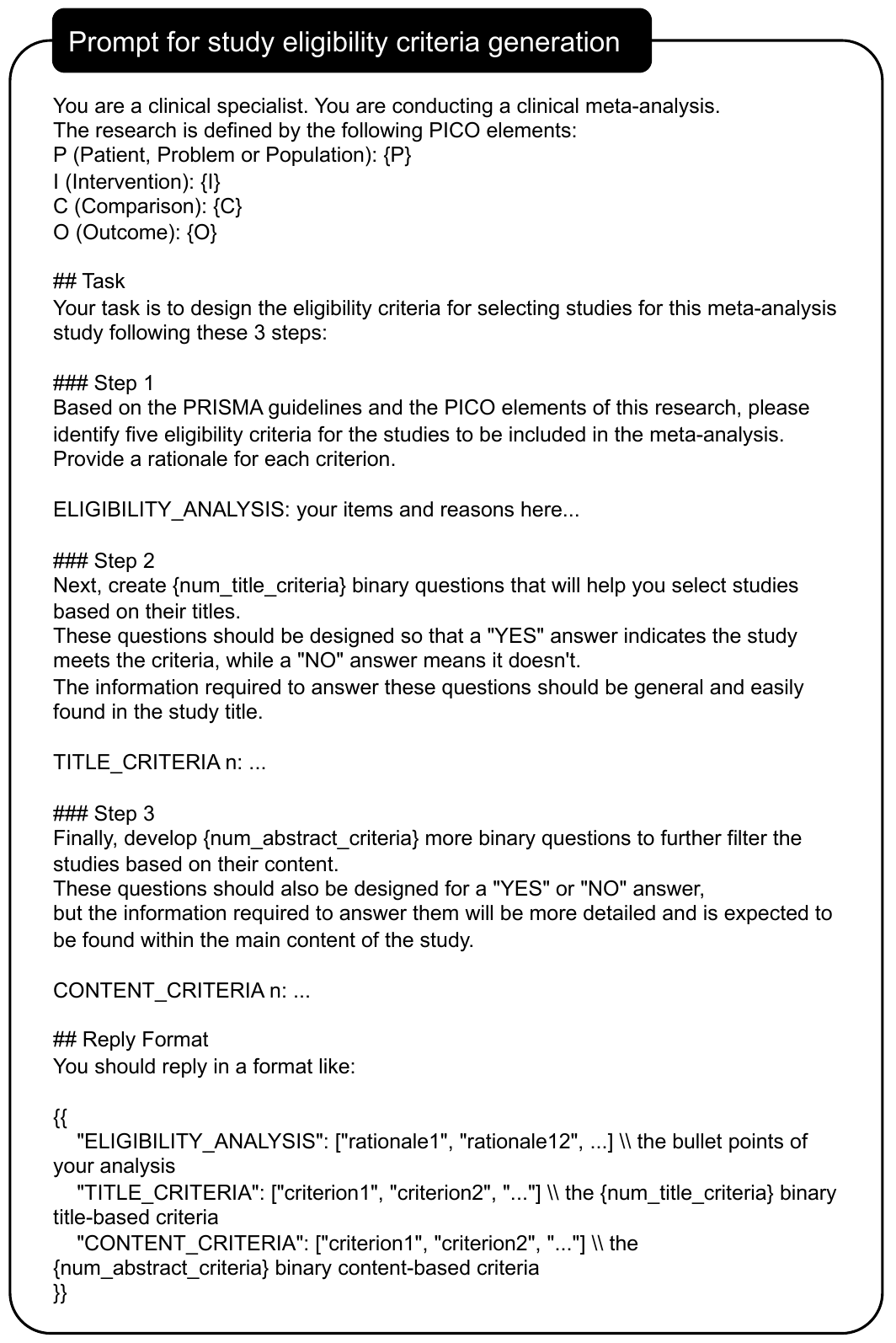}
    \caption{Prompt for study eligibility criteria generation in the literature screen.}
    \label{fig:prompt_study_eligibility_generation}
\end{figure}

\begin{figure}[!htbp]
    \centering
    \includegraphics[width=0.9\linewidth]{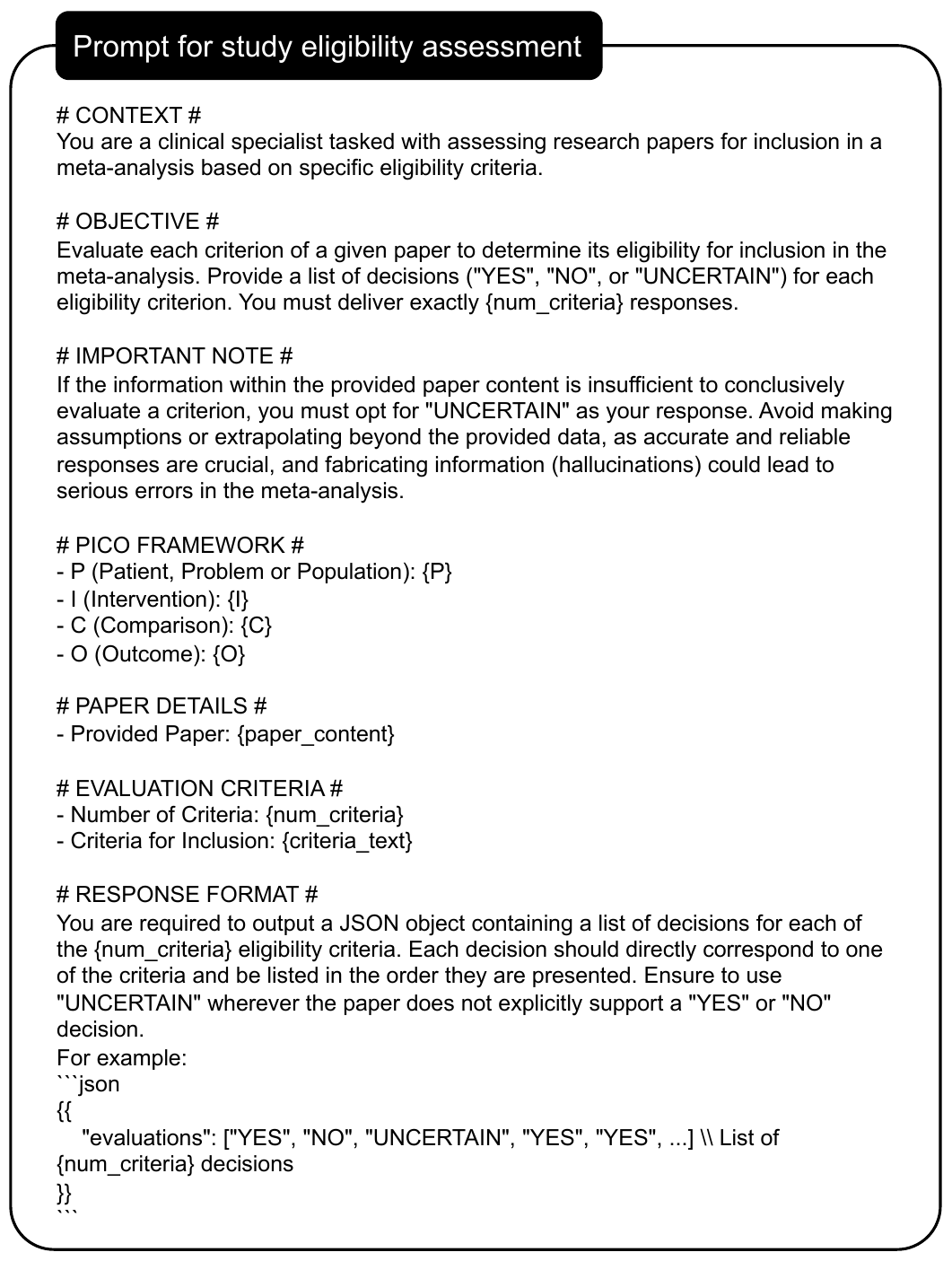}
    \caption{Prompt for study eligibility assessment in the literature screen.}
    \label{fig:prompt_eligibility_assessment}
\end{figure}

\begin{figure}[!htbp]
    \centering
    \includegraphics[width=0.9\linewidth]{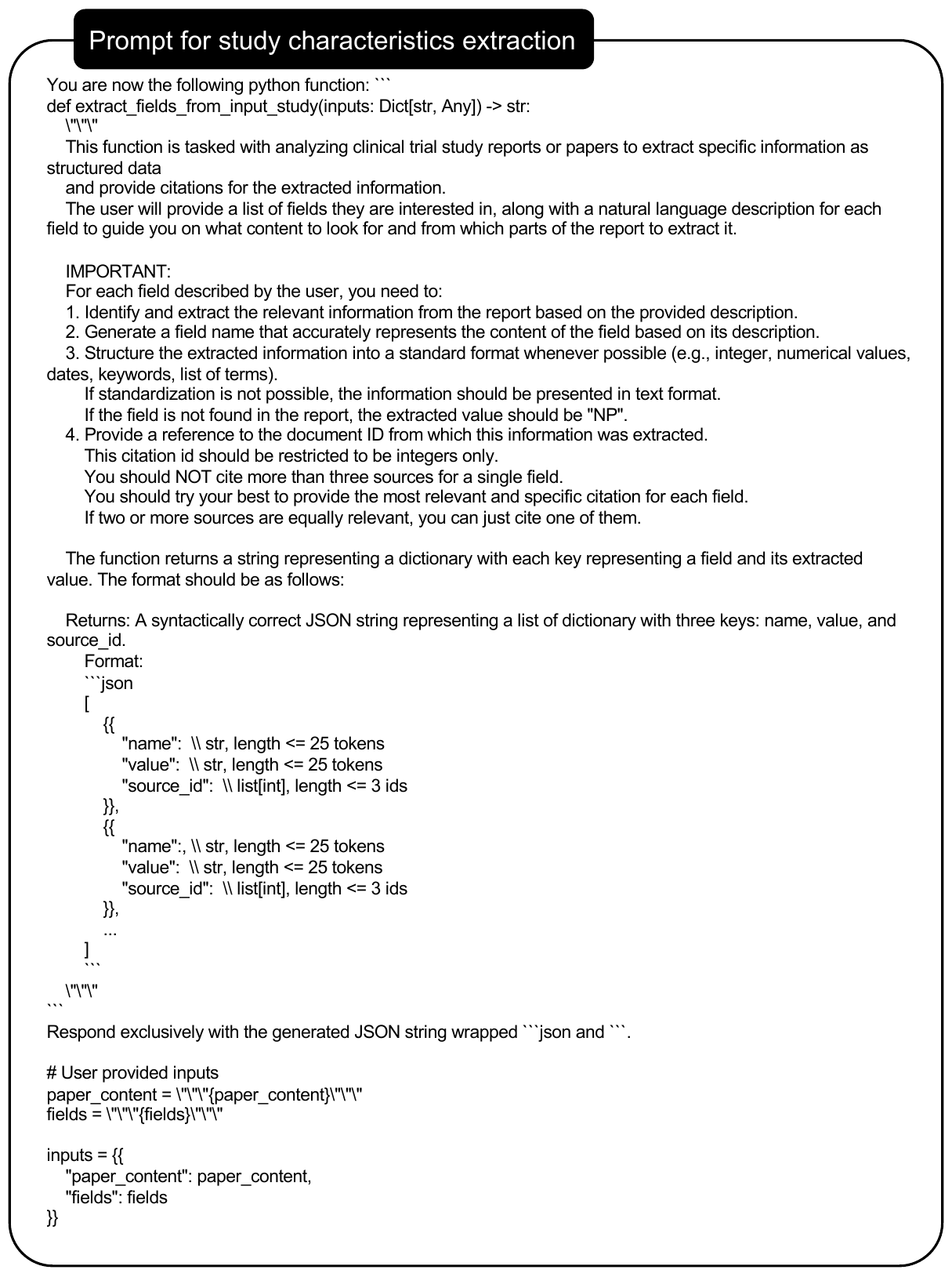}
    \caption{Prompt for study characteristics extraction in the data extraction.}
\label{fig:prompt_study_extraction}
\end{figure}

\begin{figure}[!htbp]
    \centering
\includegraphics[width=0.95\linewidth]{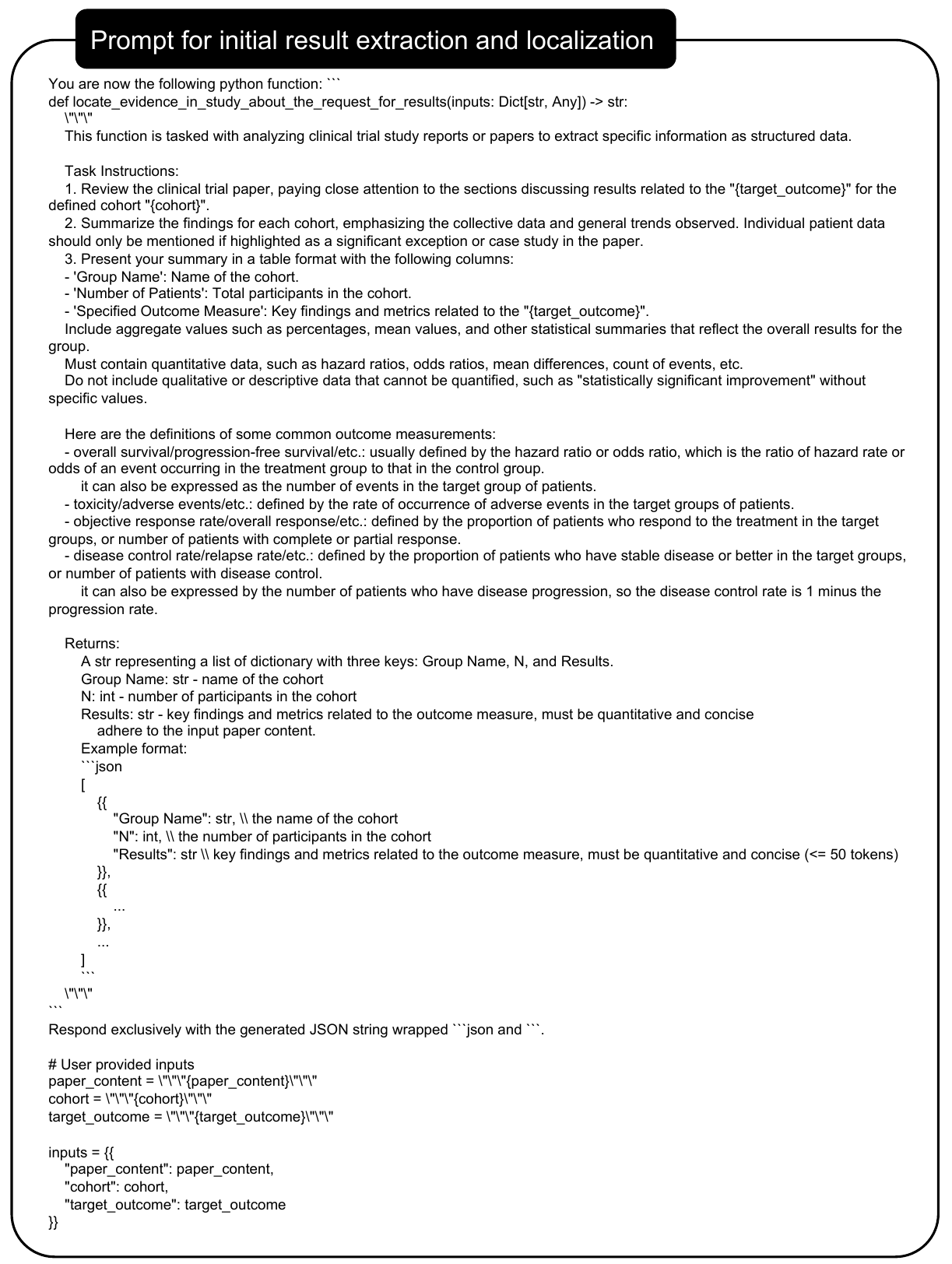}
    \caption{Prompt for the initial result extraction in the evidence synthesis.}
\label{fig:prompt_initial_result_extraction}
\end{figure}

\begin{figure}[!htbp]
    \centering
\includegraphics[width=0.95\linewidth]{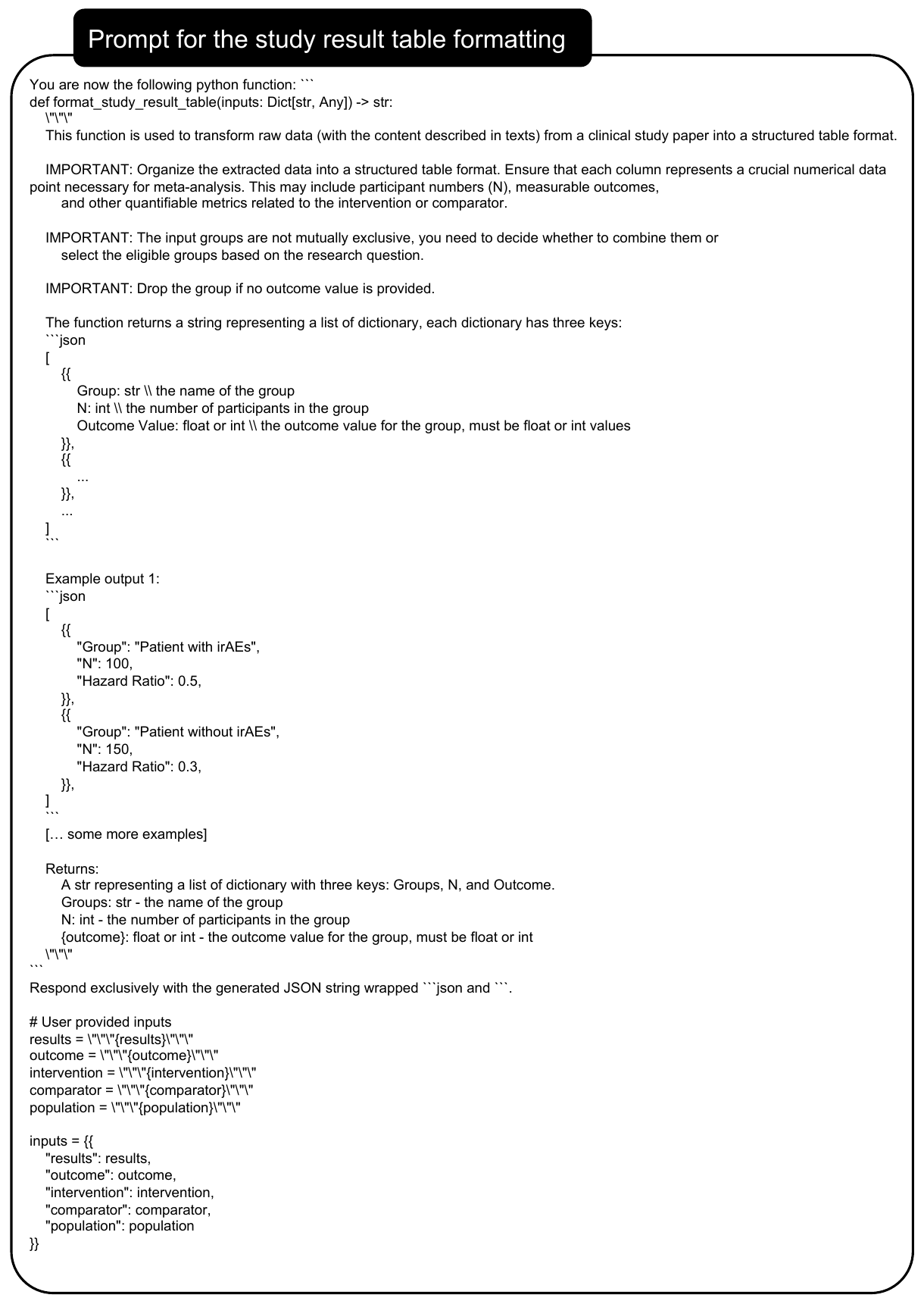}
    \caption{Prompt for the result formatting in the evidence synthesis.}
\label{fig:prompt_result_formatting}
\end{figure}

\begin{figure}[!htbp]
    \centering
\includegraphics[width=0.9\linewidth]{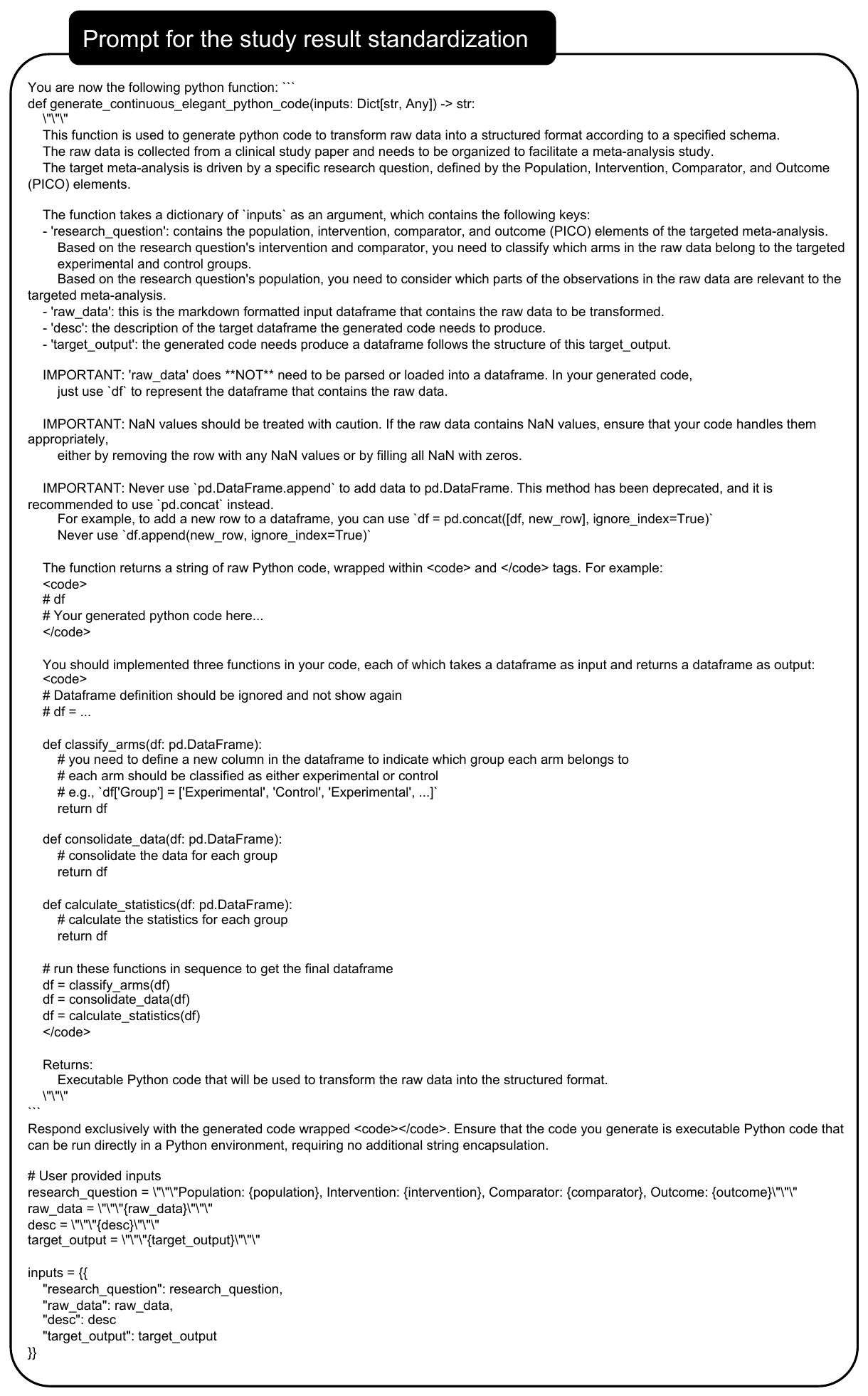}
    \caption{Prompt for the result standardization in the evidence synthesis.}
\label{fig:prompt_result_standardization}
\end{figure}

\clearpage

\begingroup
\setlength{\tabcolsep}{3pt}
\setlength\LTcapwidth{\textwidth} 

\setlength\LTleft{0pt}            
\setlength\LTright{0pt}           

\tiny

\captionsetup{font=normalsize}

\endgroup

\end{appendices}

\clearpage








\bibliographystyle{naturemag}
\bibliography{main}

\begin{thebibliography}{10}
\expandafter\ifx\csname url\endcsname\relax
  \def\url#1{\texttt{#1}}\fi
\expandafter\ifx\csname urlprefix\endcsname\relax\def\urlprefix{URL }\fi
\providecommand{\bibinfo}[2]{#2}
\providecommand{\eprint}[2][]{\url{#2}}

\bibitem{elliott2021decision}
\bibinfo{author}{Elliott, J.} \emph{et~al.}
\newblock \bibinfo{title}{Decision makers need constantly updated evidence synthesis}.
\newblock \emph{\bibinfo{journal}{Nature}} \textbf{\bibinfo{volume}{600}}, \bibinfo{pages}{383--385} (\bibinfo{year}{2021}).

\bibitem{field2010meta}
\bibinfo{author}{Field, A.~P.} \& \bibinfo{author}{Gillett, R.}
\newblock \bibinfo{title}{How to do a meta-analysis}.
\newblock \emph{\bibinfo{journal}{British Journal of Mathematical and Statistical Psychology}} \textbf{\bibinfo{volume}{63}}, \bibinfo{pages}{665--694} (\bibinfo{year}{2010}).

\bibitem{concato2017randomized}
\bibinfo{author}{Concato, J.}, \bibinfo{author}{Shah, N.} \& \bibinfo{author}{Horwitz, R.~I.}
\newblock \bibinfo{title}{Randomized, controlled trials, observational studies, and the hierarchy of research designs}.
\newblock In \emph{\bibinfo{booktitle}{Research Ethics}}, \bibinfo{pages}{207--212} (\bibinfo{publisher}{Routledge}, \bibinfo{year}{2017}).

\bibitem{borah2017analysis}
\bibinfo{author}{Borah, R.}, \bibinfo{author}{Brown, A.~W.}, \bibinfo{author}{Capers, P.~L.} \& \bibinfo{author}{Kaiser, K.~A.}
\newblock \bibinfo{title}{Analysis of the time and workers needed to conduct systematic reviews of medical interventions using data from the prospero registry}.
\newblock \emph{\bibinfo{journal}{BMJ open}} \textbf{\bibinfo{volume}{7}}, \bibinfo{pages}{e012545} (\bibinfo{year}{2017}).

\bibitem{hoffmeyer2021most}
\bibinfo{author}{Hoffmeyer, B.~D.}, \bibinfo{author}{Andersen, M.~Z.}, \bibinfo{author}{Fonnes, S.} \& \bibinfo{author}{Rosenberg, J.}
\newblock \bibinfo{title}{Most cochrane reviews have not been updated for more than 5 years.}
\newblock \emph{\bibinfo{journal}{Journal of evidence-based medicine}} \textbf{\bibinfo{volume}{14}}, \bibinfo{pages}{181--184} (\bibinfo{year}{2021}).

\bibitem{pubmednum}
\bibinfo{title}{Medline pubmed production statistics}.
\newblock \bibinfo{howpublished}{\url{https://www.nlm.nih.gov/bsd/medline_pubmed_production_stats.html}}.
\newblock \bibinfo{note}{Accessed: 2024-09-11}.

\bibitem{marshall2019toward}
\bibinfo{author}{Marshall, I.~J.} \& \bibinfo{author}{Wallace, B.~C.}
\newblock \bibinfo{title}{Toward systematic review automation: a practical guide to using machine learning tools in research synthesis}.
\newblock \emph{\bibinfo{journal}{Systematic reviews}} \textbf{\bibinfo{volume}{8}}, \bibinfo{pages}{1--10} (\bibinfo{year}{2019}).

\bibitem{brown2020language}
\bibinfo{author}{Brown, T.} \emph{et~al.}
\newblock \bibinfo{title}{Language models are few-shot learners}.
\newblock \emph{\bibinfo{journal}{Advances in Neural Information Processing Systems}} \textbf{\bibinfo{volume}{33}}, \bibinfo{pages}{1877--1901} (\bibinfo{year}{2020}).

\bibitem{wang2023can}
\bibinfo{author}{Wang, S.}, \bibinfo{author}{Scells, H.}, \bibinfo{author}{Koopman, B.} \& \bibinfo{author}{Zuccon, G.}
\newblock \bibinfo{title}{Can chatgpt write a good boolean query for systematic review literature search?}
\newblock In \emph{\bibinfo{booktitle}{Proceedings of the 46th International ACM SIGIR Conference on Research and Development in Information Retrieval}}, \bibinfo{pages}{1426--1436} (\bibinfo{year}{2023}).

\bibitem{adam2024literature}
\bibinfo{author}{Adam, G.~P.} \emph{et~al.}
\newblock \bibinfo{title}{Literature search sandbox: a large language model that generates search queries for systematic reviews}.
\newblock \emph{\bibinfo{journal}{JAMIA open}} \textbf{\bibinfo{volume}{7}}, \bibinfo{pages}{ooae098} (\bibinfo{year}{2024}).

\bibitem{wadhwa2023jointly}
\bibinfo{author}{Wadhwa, S.}, \bibinfo{author}{DeYoung, J.}, \bibinfo{author}{Nye, B.}, \bibinfo{author}{Amir, S.} \& \bibinfo{author}{Wallace, B.~C.}
\newblock \bibinfo{title}{Jointly extracting interventions, outcomes, and findings from rct reports with llms}.
\newblock In \emph{\bibinfo{booktitle}{Machine Learning for Healthcare Conference}}, \bibinfo{pages}{754--771} (\bibinfo{organization}{PMLR}, \bibinfo{year}{2023}).

\bibitem{zhang2024span}
\bibinfo{author}{Zhang, G.} \emph{et~al.}
\newblock \bibinfo{title}{A span-based model for extracting overlapping pico entities from randomized controlled trial publications}.
\newblock \emph{\bibinfo{journal}{Journal of the American Medical Informatics Association}} \textbf{\bibinfo{volume}{31}}, \bibinfo{pages}{1163--1171} (\bibinfo{year}{2024}).

\bibitem{syriani2023assessing}
\bibinfo{author}{Syriani, E.}, \bibinfo{author}{David, I.} \& \bibinfo{author}{Kumar, G.}
\newblock \bibinfo{title}{Assessing the ability of chatgpt to screen articles for systematic reviews}.
\newblock \emph{\bibinfo{journal}{arXiv preprint arXiv:2307.06464}}  (\bibinfo{year}{2023}).

\bibitem{shaib2023summarizing}
\bibinfo{author}{Shaib, C.} \emph{et~al.}
\newblock \bibinfo{title}{Summarizing, simplifying, and synthesizing medical evidence using gpt-3 (with varying success)}.
\newblock In \emph{\bibinfo{booktitle}{The 61st Annual Meeting Of The Association For Computational Linguistics}} (\bibinfo{year}{2023}).

\bibitem{wallace2021generating}
\bibinfo{author}{Wallace, B.~C.}, \bibinfo{author}{Saha, S.}, \bibinfo{author}{Soboczenski, F.} \& \bibinfo{author}{Marshall, I.~J.}
\newblock \bibinfo{title}{Generating (factual?) narrative summaries of {RCTs}: Experiments with neural multi-document summarization}.
\newblock \emph{\bibinfo{journal}{AMIA Summits on Translational Science Proceedings}} \textbf{\bibinfo{volume}{2021}}, \bibinfo{pages}{605} (\bibinfo{year}{2021}).

\bibitem{zhang2024closing}
\bibinfo{author}{Zhang, G.} \emph{et~al.}
\newblock \bibinfo{title}{Closing the gap between open source and commercial large language models for medical evidence summarization}.
\newblock \emph{\bibinfo{journal}{npj Digital Medicine}} \textbf{\bibinfo{volume}{7}}, \bibinfo{pages}{239} (\bibinfo{year}{2024}).

\bibitem{peng2023ai}
\bibinfo{author}{Peng, Y.}, \bibinfo{author}{Rousseau, J.~F.}, \bibinfo{author}{Shortliffe, E.~H.} \& \bibinfo{author}{Weng, C.}
\newblock \bibinfo{title}{Ai-generated text may have a role in evidence-based medicine}.
\newblock \emph{\bibinfo{journal}{Nature Medicine}} \textbf{\bibinfo{volume}{29}}, \bibinfo{pages}{1593--1594} (\bibinfo{year}{2023}).

\bibitem{christopoulou2023towards}
\bibinfo{author}{Christopoulou, S.~C.}
\newblock \bibinfo{title}{Towards automated meta-analysis of clinical trials: An overview}.
\newblock \emph{\bibinfo{journal}{BioMedInformatics}} \textbf{\bibinfo{volume}{3}}, \bibinfo{pages}{115--140} (\bibinfo{year}{2023}).

\bibitem{openai2024gpt4}
\bibinfo{author}{OpenAI}.
\newblock \bibinfo{title}{Gpt-4 technical report} (\bibinfo{year}{2024}).
\newblock \eprint{2303.08774}.

\bibitem{yun2023appraising}
\bibinfo{author}{Yun, H.}, \bibinfo{author}{Marshall, I.}, \bibinfo{author}{Trikalinos, T.} \& \bibinfo{author}{Wallace, B.~C.}
\newblock \bibinfo{title}{Appraising the potential uses and harms of llms for medical systematic reviews}.
\newblock In \emph{\bibinfo{booktitle}{Proceedings of the 2023 Conference on Empirical Methods in Natural Language Processing}}, \bibinfo{pages}{10122--10139} (\bibinfo{year}{2023}).

\bibitem{page2021prisma}
\bibinfo{author}{Page, M.~J.} \emph{et~al.}
\newblock \bibinfo{title}{The prisma 2020 statement: an updated guideline for reporting systematic reviews}.
\newblock \emph{\bibinfo{journal}{Bmj}} \textbf{\bibinfo{volume}{372}} (\bibinfo{year}{2021}).

\bibitem{ncicancer}
\bibinfo{author}{National\hspace{2pt}Cancer\hspace{2pt}Institute}.
\newblock \bibinfo{title}{Types of cancer treatment}.
\newblock \bibinfo{howpublished}{\url{https://www.cancer.gov/about-cancer/treatment/types}}.
\newblock \bibinfo{note}{Accessed: 2024-04-24}.

\bibitem{wu2022ai}
\bibinfo{author}{Wu, T.}, \bibinfo{author}{Terry, M.} \& \bibinfo{author}{Cai, C.~J.}
\newblock \bibinfo{title}{Ai chains: Transparent and controllable human-ai interaction by chaining large language model prompts}.
\newblock In \emph{\bibinfo{booktitle}{Proceedings of the 2022 CHI conference on human factors in computing systems}}, \bibinfo{pages}{1--22} (\bibinfo{year}{2022}).

\bibitem{bodenreider2004unified}
\bibinfo{author}{Bodenreider, O.}
\newblock \bibinfo{title}{The unified medical language system (umls): integrating biomedical terminology}.
\newblock \emph{\bibinfo{journal}{Nucleic acids research}} \textbf{\bibinfo{volume}{32}}, \bibinfo{pages}{D267--D270} (\bibinfo{year}{2004}).

\bibitem{song2020mpnet}
\bibinfo{author}{Song, K.}, \bibinfo{author}{Tan, X.}, \bibinfo{author}{Qin, T.}, \bibinfo{author}{Lu, J.} \& \bibinfo{author}{Liu, T.-Y.}
\newblock \bibinfo{title}{Mpnet: Masked and permuted pre-training for language understanding} (\bibinfo{year}{2020}).
\newblock \eprint{2004.09297}.

\bibitem{jin2023medcpt}
\bibinfo{author}{Jin, Q.} \emph{et~al.}
\newblock \bibinfo{title}{Medcpt: Contrastive pre-trained transformers with large-scale pubmed search logs for zero-shot biomedical information retrieval}.
\newblock \emph{\bibinfo{journal}{Bioinformatics}} \textbf{\bibinfo{volume}{39}}, \bibinfo{pages}{btad651} (\bibinfo{year}{2023}).

\bibitem{deeks2010statistical}
\bibinfo{author}{Deeks, J.~J.} \& \bibinfo{author}{Higgins, J.~P.}
\newblock \bibinfo{title}{Statistical algorithms in review manager 5}.
\newblock \emph{\bibinfo{journal}{Statistical methods group of the Cochrane Collaboration}} \textbf{\bibinfo{volume}{1}} (\bibinfo{year}{2010}).

\bibitem{shekelle2001validity}
\bibinfo{author}{Shekelle, P.~G.} \emph{et~al.}
\newblock \bibinfo{title}{Validity of the agency for healthcare research and quality clinical practice guidelines: how quickly do guidelines become outdated?}
\newblock \emph{\bibinfo{journal}{JAMA}} \textbf{\bibinfo{volume}{286}}, \bibinfo{pages}{1461--1467} (\bibinfo{year}{2001}).

\bibitem{hutson2024ai}
\bibinfo{author}{Hutson, M.}
\newblock \bibinfo{title}{How ai is being used to accelerate clinical trials.}
\newblock \emph{\bibinfo{journal}{Nature}} \textbf{\bibinfo{volume}{627}}, \bibinfo{pages}{S2--S5} (\bibinfo{year}{2024}).

\bibitem{wang2023pytrial}
\bibinfo{author}{Wang, Z.}, \bibinfo{author}{Theodorou, B.}, \bibinfo{author}{Fu, T.}, \bibinfo{author}{Xiao, C.} \& \bibinfo{author}{Sun, J.}
\newblock \bibinfo{title}{Pytrial: Machine learning software and benchmark for clinical trial applications}.
\newblock \emph{\bibinfo{journal}{arXiv preprint arXiv:2306.04018}}  (\bibinfo{year}{2023}).

\bibitem{jin2024pubmed}
\bibinfo{author}{Jin, Q.}, \bibinfo{author}{Leaman, R.} \& \bibinfo{author}{Lu, Z.}
\newblock \bibinfo{title}{Pubmed and beyond: biomedical literature search in the age of artificial intelligence}.
\newblock \emph{\bibinfo{journal}{Ebiomedicine}} \textbf{\bibinfo{volume}{100}} (\bibinfo{year}{2024}).

\bibitem{scells2017test}
\bibinfo{author}{Scells, H.} \emph{et~al.}
\newblock \bibinfo{title}{A test collection for evaluating retrieval of studies for inclusion in systematic reviews}.
\newblock In \emph{\bibinfo{booktitle}{Proceedings of the 40th International ACM SIGIR Conference on Research and Development in Information Retrieval}}, \bibinfo{pages}{1237--1240} (\bibinfo{year}{2017}).

\bibitem{wallace2010semi}
\bibinfo{author}{Wallace, B.~C.}, \bibinfo{author}{Trikalinos, T.~A.}, \bibinfo{author}{Lau, J.}, \bibinfo{author}{Brodley, C.} \& \bibinfo{author}{Schmid, C.~H.}
\newblock \bibinfo{title}{Semi-automated screening of biomedical citations for systematic reviews}.
\newblock \emph{\bibinfo{journal}{BMC Bioinformatics}} \textbf{\bibinfo{volume}{11}}, \bibinfo{pages}{1--11} (\bibinfo{year}{2010}).

\bibitem{kanoulas2018clef}
\bibinfo{author}{Kanoulas, E.}, \bibinfo{author}{Li, D.}, \bibinfo{author}{Azzopardi, L.} \& \bibinfo{author}{Spijker, R.}
\newblock \bibinfo{title}{Clef 2018 technologically assisted reviews in empirical medicine overview}.
\newblock In \emph{\bibinfo{booktitle}{CEUR workshop proceedings}}, vol. \bibinfo{volume}{2125} (\bibinfo{year}{2018}).

\bibitem{trikalinos2019large}
\bibinfo{author}{Trikalinos, T.} \emph{et~al.}
\newblock \bibinfo{title}{Large scale empirical evaluation of machine learning for semi-automating citation screening in systematic reviews}.
\newblock In \emph{\bibinfo{booktitle}{41st Annual Meeting of the Society for Medical Decision Making}} (\bibinfo{organization}{SMDM}, \bibinfo{year}{2019}).

\bibitem{vsuster2023automating}
\bibinfo{author}{{\v{S}}uster, S.} \emph{et~al.}
\newblock \bibinfo{title}{Automating quality assessment of medical evidence in systematic reviews: model development and validation study}.
\newblock \emph{\bibinfo{journal}{Journal of Medical Internet Research}} \textbf{\bibinfo{volume}{25}}, \bibinfo{pages}{e35568} (\bibinfo{year}{2023}).

\bibitem{yun2024automatically}
\bibinfo{author}{Yun, H.~S.}, \bibinfo{author}{Pogrebitskiy, D.}, \bibinfo{author}{Marshall, I.~J.} \& \bibinfo{author}{Wallace, B.~C.}
\newblock \bibinfo{title}{Automatically extracting numerical results from randomized controlled trials with large language models}.
\newblock \emph{\bibinfo{journal}{arXiv preprint arXiv:2405.01686}}  (\bibinfo{year}{2024}).

\bibitem{schmidt2021data}
\bibinfo{author}{Schmidt, L.} \emph{et~al.}
\newblock \bibinfo{title}{Data extraction methods for systematic review (semi) automation: Update of a living systematic review}.
\newblock \emph{\bibinfo{journal}{F1000Research}} \textbf{\bibinfo{volume}{10}} (\bibinfo{year}{2021}).

\bibitem{zhang2024leveraging}
\bibinfo{author}{Zhang, G.} \emph{et~al.}
\newblock \bibinfo{title}{Leveraging generative ai for clinical evidence synthesis needs to ensure trustworthiness}.
\newblock \emph{\bibinfo{journal}{Journal of Biomedical Informatics}} \bibinfo{pages}{104640} (\bibinfo{year}{2024}).

\bibitem{joseph2024factpico}
\bibinfo{author}{Joseph, S.~A.} \emph{et~al.}
\newblock \bibinfo{title}{Factpico: Factuality evaluation for plain language summarization of medical evidence}.
\newblock \emph{\bibinfo{journal}{arXiv preprint arXiv:2402.11456}}  (\bibinfo{year}{2024}).

\bibitem{ramprasad2023automatically}
\bibinfo{author}{Ramprasad, S.}, \bibinfo{author}{Mcinerney, J.}, \bibinfo{author}{Marshall, I.} \& \bibinfo{author}{Wallace, B.~C.}
\newblock \bibinfo{title}{Automatically summarizing evidence from clinical trials: A prototype highlighting current challenges}.
\newblock In \emph{\bibinfo{booktitle}{Proceedings of the 17th Conference of the European Chapter of the Association for Computational Linguistics: System Demonstrations}}, \bibinfo{pages}{236--247} (\bibinfo{year}{2023}).

\bibitem{chelli2024hallucination}
\bibinfo{author}{Chelli, M.} \emph{et~al.}
\newblock \bibinfo{title}{Hallucination rates and reference accuracy of chatgpt and bard for systematic reviews: Comparative analysis}.
\newblock \emph{\bibinfo{journal}{Journal of Medical Internet Research}} \textbf{\bibinfo{volume}{26}}, \bibinfo{pages}{e53164} (\bibinfo{year}{2024}).

\bibitem{spillias2023human}
\bibinfo{author}{Spillias, S.} \emph{et~al.}
\newblock \bibinfo{title}{Human-ai collaboration to identify literature for evidence synthesis}  (\bibinfo{year}{2023}).

\bibitem{lewis2020retrieval}
\bibinfo{author}{Lewis, P.} \emph{et~al.}
\newblock \bibinfo{title}{Retrieval-augmented generation for knowledge-intensive nlp tasks}.
\newblock \emph{\bibinfo{journal}{Advances in Neural Information Processing Systems}} \textbf{\bibinfo{volume}{33}}, \bibinfo{pages}{9459--9474} (\bibinfo{year}{2020}).

\bibitem{wei2022chain}
\bibinfo{author}{Wei, J.} \emph{et~al.}
\newblock \bibinfo{title}{Chain-of-thought prompting elicits reasoning in large language models}.
\newblock \emph{\bibinfo{journal}{Advances in Neural Information Processing Systems}} \textbf{\bibinfo{volume}{35}}, \bibinfo{pages}{24824--24837} (\bibinfo{year}{2022}).

\bibitem{anthropic2023claude}
\bibinfo{author}{Anthropic}.
\newblock \bibinfo{title}{Introducing the claude 3 family}.
\newblock \bibinfo{howpublished}{\url{https://www.anthropic.com/news/claude-3-family}} (\bibinfo{year}{2023}).
\newblock \bibinfo{note}{Accessed: 2024-04-24}.

\bibitem{ncbi2002biotech}
\bibinfo{author}{{National Center for Biotechnology Information (NCBI)}}.
\newblock \bibinfo{title}{Entrez programming utilities help}.
\newblock \bibinfo{howpublished}{\url{https://www.ncbi.nlm.nih.gov/books/NBK25501/}} (\bibinfo{year}{2008}).
\newblock \bibinfo{note}{Accessed: 2024-04-24}.

\end{thebibliography}

\end{document}